\documentclass[10pt,twocolumn,letterpaper]{article}

\usepackage[pagenumbers]{cvpr} 

\definecolor{myBaseBlue}{HTML}{90A4C5}
\definecolor{myRefinePeach}{HTML}{F5C8B5}

\colorlet{myBaseBlueLight}{myBaseBlue!70!white}    
\colorlet{myRefinePeachLight}{myRefinePeach!70!white} 

\usepackage{booktabs}   
\usepackage{multirow}   
\usepackage{makecell}   
\usepackage{enumitem}
\usepackage{graphicx}
\usepackage[accsupp]{axessibility}
\graphicspath{{figures/}}
\usepackage[table]{xcolor}
\usepackage{pgfplots}
\usepackage{tikz}
\pgfplotsset{compat=1.18} 
\usetikzlibrary{positioning} 
\usetikzlibrary{patterns}
\usepackage{subcaption}

\definecolor{cvprblue}{rgb}{0.21,0.49,0.74}
\usepackage[pagebackref,breaklinks,colorlinks,allcolors=cvprblue]{hyperref}

\def\confName{CVPR}
\def\confYear{2026}

\title{ReCALL: Recalibrating Capability Degradation for \\
\mbox{MLLM-based} Composed Image Retrieval}

\author{
Tianyu Yang\textsuperscript{1,2},
Chenwei He\textsuperscript{3},
Xiangzhao Hao\textsuperscript{1,2},
Tianyue Wang\textsuperscript{1,2},
Jiarui Guo\textsuperscript{4},\\
Haiyun Guo\textsuperscript{1,2,$\dagger$},
Leigang Qu\textsuperscript{5,$\dagger$},
Tat-Seng Chua\textsuperscript{5},
Jinqiao Wang\textsuperscript{1,2,6,7}\\
\textsuperscript{1}Foundation Model Research Center, Institute of Automation, Chinese Academy of Sciences \\
\textsuperscript{2}School of Artificial Intelligence, University of Chinese Academy of Sciences \\
\textsuperscript{3}Southeast University \quad
\textsuperscript{4}Beijing University of Posts and Telecommunications \\
\textsuperscript{5}National University of Singapore \quad
\textsuperscript{6}Wuhan AI Research\\
\textsuperscript{7}Guangdong Provincial Key Laboratory of Intellectual Property and Big Data,\\
Guangdong Polytechnic Normal University\\
{\tt\small \{yangtianyu2024, haoxiangzhao2023\}@ia.ac.cn, hechenwei@seu.edu.cn}\\
}

\begin{document}
\maketitle
\let\thefootnote\relax\footnotetext{$^\dagger$ Corresponding author.}
\begin{abstract}

Composed Image Retrieval (CIR) aims to retrieve target images based on a hybrid query comprising a reference image and a modification text. Early dual-tower Vision–Language Models (VLMs) struggle with cross-modality compositional reasoning required for this task. While adapting generative Multimodal Large Language Models (MLLMs) for retrieval offers a promising direction, we identify that this strategy overlooks a fundamental issue: compressing a generative MLLM into a single-embedding discriminative retriever triggers a paradigm conflict, which leads to \textbf{Capability Degradation}—the deterioration of native fine-grained reasoning after retrieval adaptation. To address this challenge, we propose \textbf{ReCALL}, a model-agnostic framework that follows a \emph{diagnose–generate–refine} pipeline: First, we diagnose cognitive blind spots of the retriever via self-guided informative instance mining. Next, we generate corrective instructions and triplets by prompting the foundation MLLM and conduct quality control with VQA-based consistency filtering. Finally, we refine the retriever through continual training on these triplets with a grouped contrastive scheme, thereby internalizing fine-grained visual–semantic distinctions and realigning the discriminative embedding space of retriever with intrinsic compositional reasoning within the MLLM. Extensive experiments on CIRR and FashionIQ show that ReCALL consistently recalibrates degraded capabilities and achieves state-of-the-art performance. Code is available at \href{https://github.com/RemRico/Recall}{https://github.com/RemRico/Recall}.
\end{abstract}    
\section{Introduction}
\label{sec:intro}

Composed Image Retrieval (CIR) retrieves a target image given a composed query that combines a reference image and a textual modification. Due to the vast application potential in domains such as e-commerce and design, it has attracted a surge of research interests~\cite{vo2019composing,wen2023target,baldrati2022effective,qu2025vincie} recently, enabling users to articulate more complex and precise search intent compared with traditional image retrieval~\cite{10.1007/978-3-319-46466-4_15,anwaar2021compositional,pham2024composing,yang2022vision,hao2025referring}.

 \begin{figure*}[t]
  \centering
  \includegraphics[width=0.95\textwidth]{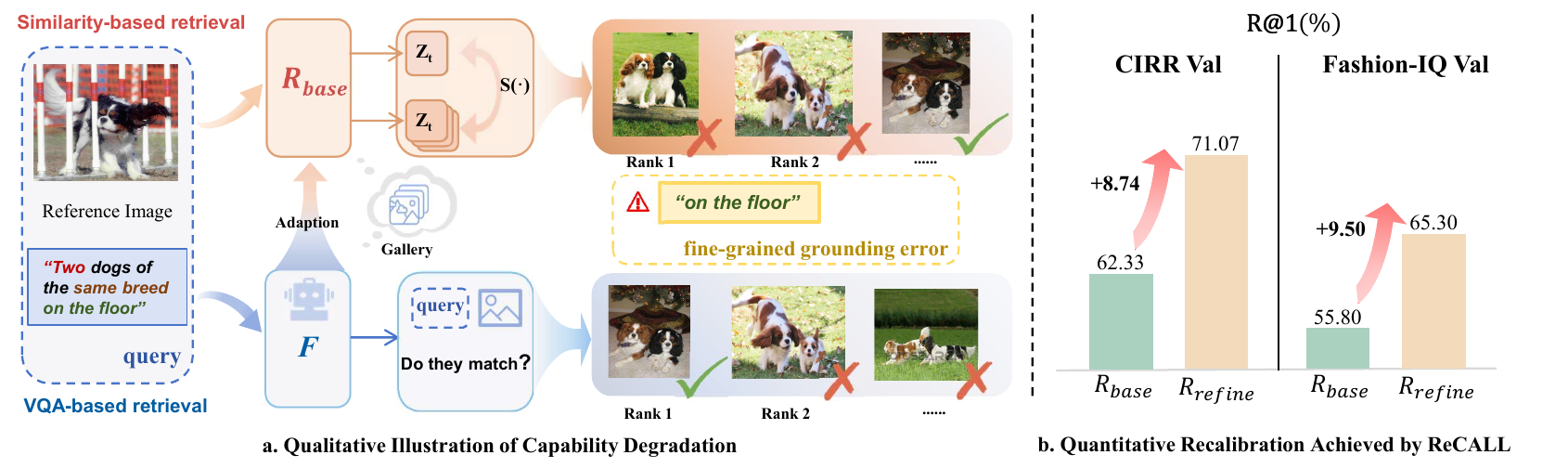}
  \caption{Empirical illustration of Capability Degradation and the effectiveness of ReCALL ($\mathcal{R}_{\text{refine}}$). (a) We compare the Foundation MLLM ($\mathcal{F}$) under its native VQA-based generative paradigm with its fine-tuned retrieval counterpart ($\mathcal{R}_{\text{base}}$) under a similarity-based discriminative paradigm using a challenging query that requires fine-grained reasoning. The base retriever $\mathcal{R}_{\text{base}}$ fails due to fine-grained grounding errors, while $\mathcal{F}$ succeeds through step-wise reasoning. (b) Quantitative evidence of Capability Degradation and Recalibration. We test $\mathcal{R}_{\text{base}}$ on a subset of 1k instances where $\mathcal{F}$ successfully retrieves the target (i.e., $\mathcal{F}$ achieves 100\% R@1). The low R@1 performance of $\mathcal{R}_{\text{base}}$ (only 62.33\% on CIRR and 55.80\% on FashionIQ) on this $\mathcal{F}$-solvable subset provides quantifiable proof of capability degradation. Our proposed ReCALL framework effectively recovers the lost abilities, elevating $\mathcal{R}_{\text{base}}$ to $\mathcal{R}_{\text{refine}}$ with significant gains.}
  \label{fig:intro2}
\end{figure*}

Early dual-tower vision–language models (VLMs)~\cite{liu2021image,levy2024data,bai2023sentence,anwaar2021compositional,baldrati2022effective,qu2021dynamic} struggle with fine-grained compositional reasoning because of shallow cross-modal alignment and limited modality interaction. In contrast, Multimodal Large Language Models (MLLMs) ~\cite{bai2023qwen,wang2024qwen2,bai2025qwen2,lu2024deepseek,zhu2025internvl3,xu2024llava}, benefiting from deep-fusion architectures and robust instruction-following abilities, are naturally suited for CIR. Recent works therefore adapt MLLMs to retrieval via contrastive learning~\cite{liu2025lamra,Jiang2025VLM2Vec,Lin2025MMEmbed,qu2024tiger,hao2025referring,hao2026trace}. Despite the remarkable progress, we identify a critical and overlooked challenge: adapting the MLLM’s native generative paradigm (focusing on step-wise reasoning) into a single-embedding discriminative paradigm (highlighting on vector similarity) introduces an intrinsic paradigm conflict, fundamentally degrading the model’s compositional reasoning capabilities, particularly in fine-grained grounding and relational understanding.

To substantiate the Capability Degradation phenomenon, we conduct qualitative and quantitative analyses comparing the Foundation MLLM ($\mathcal{F}$) in its native generative mode with its fine-tuned retrieval counterpart ($\mathcal{R}_{\text{base}}$). Qualitatively, as shown in Fig.~\ref{fig:intro2} (Left), $\mathcal{R}_{\text{base}}$ fails to retrieve the target for a challenging query, whereas $\mathcal{F}$ succeeds via zero-shot VQA, indicating a suppression of intrinsic compositional reasoning. Quantitatively (Fig.~\ref{fig:intro2}, Right), this degradation is unequivocal: $\mathcal{R}_{\text{base}}$ suffers a severe performance drop, achieving an R@1 of only 62.33\% and 55.80\% on the $\mathcal{F}$-solvable subsets of CIRR and FashionIQ, respectively.

To address this issue, we propose \textbf{ReCALL}, a model-agnostic framework that recalibrates degraded capabilities from the foundation model and internalizes them into the retriever’s representations. Our core idea is to leverage the MLLM’s stepwise \emph{native} reasoning signals to supervise the \emph{foreign} single-embedding retrieval space, within a \textbf{diagnose–generate–refine} pipeline. To this end, we first \textbf{diagnose} the retrieval model’s cognitive blind spots through a self-guided informative instance mining procedure, which autonomously discovers samples that the retrieval model currently struggles to distinguish. Next, we aim to \textbf{generate} corrective supervision that explicitly targets these deficiencies. Specifically, we prompt the foundation model with Chain-of-Thought (CoT)~\cite{Chu2023NavigateTE,Sun2025CoTMR,Yang2024LDRE,Tang2025GLoRE,Liu2024ICV} to generate high-quality, \emph{corrective} textual instructions for the informative instances, forming new triplets. These triplets exhibit subtle but semantically meaningful variations across both visual and textual modalities, precisely capturing the nuances that the retrieval model previously failed to distinguish. Crucially, to ensure the reliability of these generated signals, we incorporate a VQA-based consistency check to filter out noise. Finally, we \textbf{refine} the retrieval model through a novel Grouped Contrastive Learning strategy. By constructing training batches that explicitly contrast the original queries with their corrected counterparts, we encourage the model to internalize these fine-grained visual–semantic distinctions, thereby realigning its discriminative representation space with the foundation model’s intrinsic compositional reasoning capabilities.

In summary, our main contributions are as follows:
\begin{itemize}
    \item We identify a critical challenge in adapting MLLMs to CIR, termed Capability Degradation, where the native compositional reasoning of the model deteriorates during retrieval-oriented fine-tuning.
    \item We propose a model-agnostic framework, ReCALL, to recalibrate the embedding space of the retriever with the MLLM’s compositional reasoning through a \emph{diagnose-generate-refine} pipeline.
    \item Extensive experiments demonstrate that ReCALL effectively recalibrates the degraded capabilities, ultimately achieving state-of-the-art performance on mainstream CIR benchmarks, including CIRR~\cite{liu2021image} and FashionIQ~\cite{wu2021fashion}.
\end{itemize}

\section{Related Work}
\label{sec:related_work}

\subsection{Composed Image Retrieval}
CIR aims to retrieve a target image based on a hybrid-modal query. Early approaches~\cite{baldrati2022effective,Baldrati2023CIR_CLIP_TOMM,Delmas2022ARTEMIS,liu2021image} primarily follow the VLM framework (\emph{e.g.}, CLIP~\cite{Radford2021CLIP_ICML}), lacking thorough fusion between query modalities. They resort to external fusion modules~\cite{liu2021image,chen2020image,levy2024data,liu2023candidate,baldrati2022conditioned,wen2023target,Levy2023DataRA,wang2026wiser} or concatenation via pseudo-tokens~\cite{bai2023sentence,gal2022image,saito2023pic2word,Tang2023ContextI2WMI,Cohen2022Fluffy}, but are constrained by a fundamental architectural flaw, \emph{i.e.}, shallow alignment~\cite{Hsieh2023SugarCrepe,Wu2023LinguisticPriorsCompositionality}.
To overcome this limitation, recent research has shifted towards MLLMs, with CIR-LVLM~\cite{sun2025leveraging} as a representative example that leverages an LVLM as a user-intent–aware encoder for CIR. Benefiting from deep fusion and instruction-following, such adaptations have consistently demonstrated superior performance on mainstream benchmarks. Despite the remarkable progress, we argue that their adaptation for discriminative retrieval can introduce \emph{Capability Degradation}. This conflict leads to the degradation of the model’s native fine-grained reasoning, a critical gap our work aims to address.

\subsection{Self-Improvement for MLLMs}
Self-improvement has proved effective for large language models: STaR bootstraps from model-produced rationales to reinforce correct reasoning~\cite{Zelikman2022STAR}, while Reflexion and Self-Refine introduce explicit self-feedback loops to iteratively revise and correct outputs~\cite{Shinn2023Reflexion,Madaan2023SelfRefine}. In contrast, contemporary CIR adaptations of MLLMs predominantly adopt a \emph{single-stage, static fine-tuning} paradigm—fine-tuning unified encoders on curated benchmarks without online diagnosis-and-repair~\cite{sun2025leveraging,Jiang2025VLM2Vec,liu2025lamra,Lin2025MMEmbed}. To bridge this gap, ReCALL instantiates a retrieval-oriented self-improvement loop aligned with our \emph{diagnose–generate–refine} pipeline.

\section{Method}
\label{sec:method}

\begin{figure*}[t]
  \centering
  \includegraphics[width=\textwidth]{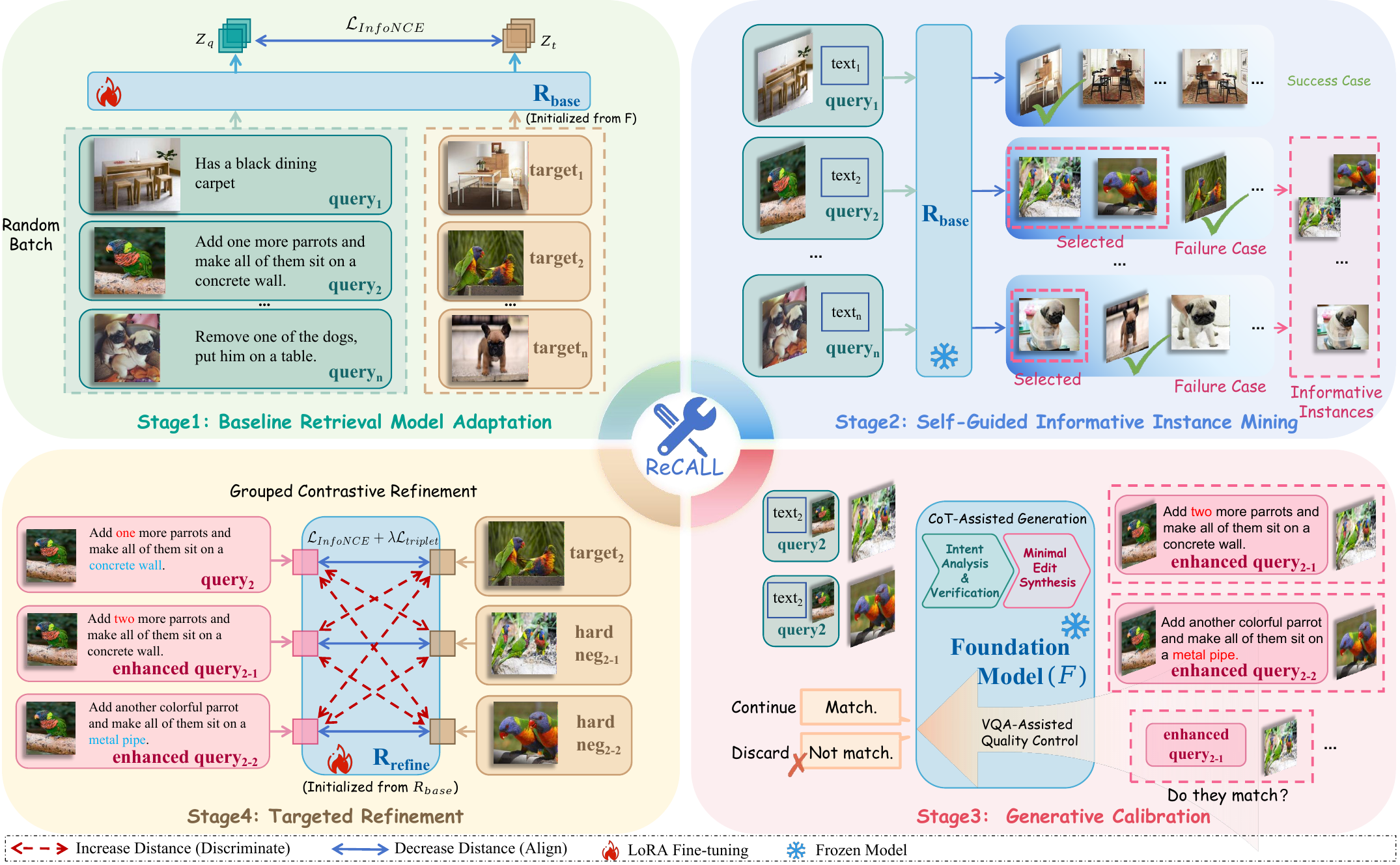}
  \caption{
  Overview of the \textbf{ReCALL} framework. 
  (1) \textbf{Stage 1:} A baseline retriever $\mathcal{R}_{base}$ is adapted from the foundation model $\mathcal{F}$ via standard fine-tuning. 
  (2) \textbf{Stage 2 (Diagnose):} $\mathcal{R}_{base}$ surfaces its own failure cases via self-guided informative instance mining. 
  (3) \textbf{Stage 3 (Generate):} Leveraging native reasoning (CoT), $\mathcal{F}$ synthesizes minimally edited corrective instructions for the mined informative instances. 
  (4) \textbf{Stage 4 (Refine):} Based on the original and enhanced triplets, a Grouped Contrastive Refinement strategy is employed to produce the final $\mathcal{R}_{refine}$, effectively recalibrating the degraded capabilities.
  }
  \label{fig:method}
\end{figure*}

This section outlines the ReCALL framework. As shown in Fig.~\ref{fig:method}, we first formalize the task and introduce the model components (Sec.~\ref{sec:formulation}), then describe the baseline adaptation procedure (Sec.~\ref{sec:stage1}). We next present the diagnose–generate–refine pipeline, including self-guided informative instance mining (Sec.~\ref{sec:stage2}), generative calibration (Sec.~\ref{sec:stage3}), and targeted refinement (Sec.~\ref{sec:stage4}).

\subsection{Problem Formulation}
\label{sec:formulation}

CIR is defined as follows: given a reference image $I_r$ and a modification text $T_m$, the goal is to retrieve the target image $I_t$ from a large gallery. We introduce the following model entities used throughout this work:
\begin{itemize}
  \item \textbf{Foundation Model ($\mathcal{F}$):} An MLLM with strong generative and reasoning capabilities, providing the intrinsic compositional reasoning that our framework leverages.
  \item \textbf{Baseline Retrieval Model ($\mathcal{R}_{base}$):} A retrieval model fine-tuned from $\mathcal{F}$ on CIR triplets using contrastive learning. While it offers basic retrieval performance, it still suffers from the capability degradation described in Sec.~\ref{sec:intro}. This model serves as the starting point for our diagnose–generate–refine pipeline.
  \item \textbf{Refined Model ($\mathcal{R}_{refine}$):} The final model variant of our framework. It addresses the capability degradation in $\mathcal{R}_{base}$ by absorbing the compositional reasoning of $\mathcal{F}$, yielding a recalibrated and more robust retriever.
\end{itemize}

\subsection{Stage 1: Baseline Retrieval Model Adaptation}
\label{sec:stage1}

The first stage adapts $\mathcal{F}$ into a retrieval model to attain basic discriminative ability, yielding the baseline retriever ($\mathcal{R}_{base}$). It provides a stable starting point for the subsequent diagnose–generate–refine pipeline.

To maximize retention of pre-trained knowledge, $\mathcal{R}_{base}$ is initialized directly from $\mathcal{F}$. We then fine-tune the model on CIR triplets $(I_r, T_m, I_t)$ via InfoNCE~\cite{oord2018representation}, encouraging the query representation $z_q$ to align with its positive target $z_t$ while pushing away in-batch negatives.

While this learning process yields a functional retriever, the resulting model inevitably suffers from capability degradation, \emph{i.e.}, discriminative fine-tuning may compromise the fine-grained compositional reasoning within $\mathcal{F}$. To address this issue, the following \emph{diagnose} stage is explicitly designed to detect and remediate these consequent blind spots.

\subsection{Stage 2: Self-Guided Informative Instance Mining}
\label{sec:stage2}

To effectively recalibrate $\mathcal{R}_{base}$, we introduce a self-guided informative instance mining strategy to probe the decision boundaries of $\mathcal{R}_{base}$ that are most susceptible to the capability degradation discussed in Sec.~\ref{sec:intro}.

First, we perform retrieval inference on the training set using $\mathcal{R}_{base}$. We exclude queries where the ground-truth $I_t$ is successfully ranked first, assuming these instances reflect sufficient discriminative power. Instead, we focus on the failure cases, as they likely harbor the most informative signals regarding where the fine-tuning process has compromised the model's original reasoning capabilities.

For each failure case, we construct a set of \emph{informative instances}, denoted as $\{I_h\}$, by isolating the \textbf{top-$K$} images erroneously ranked above the ground truth $I_t$. These instances are highly informative precisely because they share subtle visual or semantic nuances with the target, successfully deceiving the retriever due to its degraded fine-grained reasoning. Consequently, these specific instances serve as critical anchors for the subsequent calibration stage, pinpointing exactly where the model's decision boundaries require refinement.

\subsection{Stage 3: Generative Calibration}
\label{sec:stage3}
Given the informative instances $\{I_h\}$ identified in Sec.~\ref{sec:stage2}, we exploit the intrinsic generative and reasoning capabilities of $\mathcal{F}$ to synthesize corrective supervision signals. The goal is to articulate how the original instruction $T_m$ should be \emph{minimally} adjusted to align with each $I_h$ while preserving the original distribution, effectively transforming a failure case into a high-quality training example.

\vspace{4pt}
\noindent\textbf{CoT-Assisted Generation.}
In general, an informative instance $I_h$ differs from the ground-truth $I_t$ only in subtle visual aspects, as shown in Fig.~\ref{fig:method}. Such subtle differences exactly reflect the discriminative weaknesses of $\mathcal{R}_{base}$, which could be repurposed into informative supervision for continual learning. To achieve this goal, we construct minimal edits to $T_m$ to obtain $\tilde{T}_m$ that precisely reflect the visual discrepancy between $I_t$ and $I_h$, so that the new triplet $(I_r, \tilde{T}_m, I_h)$ conveys the informative supervision to further unlock the fine-grained discriminative powers of the retriever. Concretely, we employ a multi-step reasoning procedure with $\mathcal{F}$ to identify the semantic mismatch between the query $(I_r, T_m)$ and $I_h$, and then apply the necessary minimal textual changes. This procedure consists of the following two steps:

\begin{enumerate}[leftmargin=*]
\item \textbf{Intent Decomposition \& Verification:} $\mathcal{F}$ decomposes $T_m$ into atomic intents and verifies each against $(I_r, I_h)$, determining which intents are violated in $I_h$.
\item \textbf{Minimal Edit Synthesis:} $\mathcal{F}$ retains the valid intents consistent with $(I_r, I_h)$ and regenerates only the violated components, producing the corrected instruction $\tilde{T}_m$.
\end{enumerate}
This procedure induces the corrective triplet $(I_r, \tilde{T}_m, I_h)$, which provides dense and fine-grained supervision: the minimal textual edits from $T_m$ to $\tilde{T}_m$ directly mirror the subtle visual differences between $I_t$ and $I_h$, encouraging the retriever to learn from these challenging and informative distinctions explicitly.

\vspace{2pt}
\noindent\textbf{VQA-Assisted Quality Control.}
To ensure reliability, we further apply a semantic consistency check strategy with the discriminative understanding of $\mathcal{F}$. Specifically, we prompt $\mathcal{F}$ with targeted VQA questions about key attributes in $\tilde{T}_m$. Only triplets receiving high-confidence and internally consistent answers are retained for the final refinement stage.

\subsection{Stage 4: Targeted Refinement}
\label{sec:stage4}

The final stage performs targeted refinement of $\mathcal{R}_{base}$ guided by the corrective supervision generated in Sec.~\ref{sec:stage3}. We initialize $\mathcal{R}_{refine}$ from $\mathcal{R}_{base}$ and train it to internalize the fine-grained distinctions revealed by the newly constructed triplets. This is achieved through two key components: grouped contrastive refinement and a dual optimization objective.

\subsubsection{Grouped Contrastive Refinement}
To fully exploit the corrective supervision from Sec.~\ref{sec:stage3} for continual learning, we adopt a structured batching strategy. For each query, we build a \emph{micro-group} containing both the original positive triplet $(I_r, T_m, I_t)$ and its corrective counterpart $(I_r, \tilde{T}_m, I_h)$. This grouping exposes the model’s blind spots within a single gradient update. By placing $I_t$ together with its corresponding informative instance $I_h$, as well as the minimally different instructions $T_m$ and $\tilde{T}_m$ in the same batch, the model is encouraged to discriminate between visually adjacent samples via fine-grained semantic cues. As a result, these mined informative instances serve as effective anchors for refining decision boundaries.

\subsubsection{Dual Optimization Objective}
To balance global retrieval performance and fine-grained correction, we optimize $\mathcal{R}_{refine}$ using a hybrid objective:

\vspace{4pt}
\noindent\textbf{InfoNCE Loss ($\mathcal{L}_{infoNCE}$).}
We apply the standard InfoNCE loss~\cite{oord2018representation} over the entire batch, preserving the global structure learned in Sec.~\ref{sec:stage1} while accommodating new distinctions:
\begin{equation}
\mathcal{L}_{infoNCE}
= -\log
\frac{\exp(s(z_q, z_{t^+}) / \tau)}
{\sum_{z_t \in \mathcal{B}} \exp(s(z_q, z_t) / \tau)},
\end{equation}
where $\mathcal{B}$ denotes the batch of target representations, $\tau$ is the temperature parameter, and $s(u,v)=\frac{u^\top v}{\|u\|\|v\|}$ denotes the cosine similarity. Additionally, $z_q$ is the query representation derived from the input $(I_r, T_m)$, $z_{t^+}$ is the representation of the positive ground-truth image $I_t$, and $z_t$ is a generic target representation from the batch $\mathcal{B}$.

\vspace{4pt}
\noindent\textbf{In-Group Triplet Margin Loss ($\mathcal{L}_{triplet}$).}
To explicitly enforce the separation between the target and the specific informative instance within each micro-group, we add a margin-based loss~\cite{Schroff2015FaceNet}:
\begin{equation}
\mathcal{L}_{triplet} = \max(0,\; s(z_q, z_{t^-}) - s(z_q, z_{t^+}) + m),
\end{equation}
where $m$ is a margin hyperparameter, and $z_{t^-}$ corresponds to the $I_h$ identified in the diagnose stage.

Combining the above two losses, the final objective is formulated as:
\begin{equation}
\mathcal{L}_{total} = \mathcal{L}_{infoNCE} + \lambda \mathcal{L}_{triplet},
\end{equation}
where $\lambda$ balances global alignment and targeted refinement. This optimization strategy effectively counteracts capability degradation, re-incentivizing the model's fine-grained compositional reasoning.

In summary, ReCALL implements a diagnose–generate–refine pipeline that surfaces the failure cases of the baseline retriever, generates precise corrective supervision with $\mathcal{F}$, and internalizes these distinctions through targeted refinement. This process counteracts capability degradation and restores the fine-grained compositional reasoning required for reliable CIR.
\section{Experiments}
\label{sec:experiments}

\begin{table*}
  \centering
  \definecolor{dspgreen}{rgb}{0.0, 0.6, 0.0}
  \setlength{\tabcolsep}{3pt}
  \setlength{\aboverulesep}{0pt}
  \setlength{\belowrulesep}{0pt}
  \renewcommand{\arraystretch}{1.02} 
  
  \caption{Performance comparison on the CIRR test set. We compare the proposed \textbf{ReCALL} ($\mathcal{R}_{\text{refine}}$) against state-of-the-art methods. $\mathcal{R}_{\text{base}}$ denotes the baseline retriever obtained after \textbf{Stage~1}, which serves as the starting point for our refinement pipeline. The ``Avg.'' metric is computed as $(R@5 + R_{\text{subset}}@1)/2$. Best results are in \textbf{bold}, and the second-best are \underline{underlined}. The bottom row ($\Delta$) highlights the \textit{relative improvement} of ReCALL over $\mathcal{R}_{\text{base}}$, quantifying the efficacy of our recalibration strategy.}
  \label{tab:cirr}
  \small
  \begin{tabular}{l c cccc ccc c}
    \toprule
    \multirow{2}{*}{Method} & \multirow{2}{*}{Venue} & \multicolumn{4}{c}{$\text{Recall}@K$} & \multicolumn{3}{c}{$\text{Recall}_{\text{subset}}@K$} & \multirow{2}{*}{Avg.} \\
    \cmidrule(lr){3-6} \cmidrule(lr){7-9}
     & & $K=1$ & $K=5$ & $K=10$ & $K=50$ & $K=1$ & $K=2$ & $K=3$ & \\
    \midrule
    TIRG~\cite{Vo2019TIRG}       & \textcolor{gray}{CVPR'19}   & 14.61 & 48.37 & 64.08 & 90.03 & - & - & - & - \\
    ARTEMIS~\cite{Delmas2022ARTEMIS} & \textcolor{gray}{ICLR'22}   & 16.96 & 46.10 & 61.31 & 87.73 & 39.99 & 62.20 & 75.67 & 43.05 \\
    TG-CIR~\cite{wen2023target}   & \textcolor{gray}{MM'23}     & 45.25 & 78.29 & 87.16 & 97.30 & 72.84 & 89.25 & 95.13 & 75.57 \\
    SPRC~\cite{bai2023sentence}   & \textcolor{gray}{ICLR'24}   & 51.96 & 82.12 & 89.74 & 97.69 & 80.65 & 92.31 & 96.60 & 81.39 \\
    LIMN~\cite{wen2023self}       & \textcolor{gray}{TPAMI'24}  & 43.64 & 75.37 & 85.42 & 97.04 & 69.01 & 86.22 & 94.19 & 72.19 \\
    CoVR-2~\cite{ventura2024covr} & \textcolor{gray}{TPAMI'24}  & 50.43 & 81.08 & 88.89 & 98.05 & 76.75 & 90.34 & 95.78 & 79.28 \\
    CaLa~\cite{jiang2024cala}     & \textcolor{gray}{SIGIR'24}  & 49.11 & 81.21 & 89.59 & 98.00 & 76.27 & 91.04 & 96.46 & 78.74 \\
    ENCODER~\cite{li2025encoder}  & \textcolor{gray}{AAAI'25}   & 46.10 & 77.98 & 87.16 & 97.64 & 76.92 & 90.41 & 95.95 & 77.45 \\
    CIR-LVLM~\cite{sun2025leveraging} & \textcolor{gray}{AAAI'25} & \underline{53.64} & 83.76 & 90.60 & 97.93 & 79.12 & 92.33 & 96.67 & 81.44 \\
    QuRe~\cite{kwak2025qure}      & \textcolor{gray}{ICML'25}   & 52.22 & 82.53 & 90.31 & 98.17 & 78.51 & 91.28 & 96.48 & 80.52 \\
    CCIN~\cite{tian2025ccin}      & \textcolor{gray}{CVPR'25}   & 53.41 & \underline{84.05} & \underline{91.17} & 98.00 & - & - & - & - \\
    TME~\cite{li2025learning}     & \textcolor{gray}{CVPR'25}   & 53.42 & 82.99 & 90.24 & 98.15 & \underline{81.04} & \underline{92.58} & \underline{96.94} & \underline{82.01} \\
    \midrule
    Baseline ($\mathcal{R}_{\text{base}}$)          & - & 51.23 & 82.15 & 90.20 & \underline{98.20} & 77.57 & 91.83 & 96.34 & 79.86 \\
    \rowcolor{gray!15}
    \textbf{ReCALL} ($\mathcal{R}_{\text{refine}}$) & - & \textbf{55.52} & \textbf{84.07} & \textbf{91.83} & \textbf{98.55} & \textbf{81.49} & \textbf{93.35} & \textbf{97.64} & \textbf{82.81} \\
    \midrule
    \multicolumn{2}{l}{\textit{Improvement ($\Delta$)}} 
    & \textit{\textcolor{dspgreen}{+8.38\%}} 
    & \textit{\textcolor{dspgreen}{+2.34\%}} 
    & \textit{\textcolor{dspgreen}{+1.81\%}} 
    & \textit{\textcolor{dspgreen}{+0.36\%}} 
    & \textit{\textcolor{dspgreen}{+5.06\%}} 
    & \textit{\textcolor{dspgreen}{+1.65\%}} 
    & \textit{\textcolor{dspgreen}{+1.35\%}} 
    & \textit{\textcolor{dspgreen}{+3.70\%}} \\
    \bottomrule
  \end{tabular}
\end{table*}

\subsection{Datasets and Evaluation Metrics}
\label{sec:datasets}

\paragraph{Datasets.} Following prior work~\cite{tian2025ccin,Xing2025ConTextCIR,li2025learning}, we evaluate our method on two widely adopted CIR benchmarks: FashionIQ and CIRR.

\textbf{FashionIQ}~\cite{wu2021fashion} is a fine-grained benchmark dataset focusing on the fashion domain. It consists of triplets sourced from e-commerce websites, where each triplet comprises a reference image, a target image, and a natural language instruction describing the desired modifications. The dataset is divided into three categories: Dress, Shirt, and Top\&Tee, making it particularly suitable for assessing the ability of models to understand subtle attribute changes such as color, pattern, and style.

\textbf{CIRR}~\cite{liu2021image} serves as a testbed for generalization in open-domain scenarios. It is derived from the real-world NLVR2~\cite{suhr2019nlvr2} dataset, with triplets involving complex object interactions and relational changes. In contrast to the domain-specific nature of FashionIQ, it offers a complementary and challenging evaluation scenario.

\vspace{6pt}
\noindent\textbf{Evaluation Metrics.} Following standard protocol~\cite{Chen2022ComposedIR,Sun2025CoTMR,sun2025leveraging}, we adopt $\text{Recall}@K$ ($R@K$) as our primary metric, which measures the percentage of queries where the target appears in the top-$K$ results. For FashionIQ, we report $R@10$ and $R@50$ averaged across its three categories. For CIRR, we report $R@1$, $R@5$, $R@10$, and $R@50$. Additionally, for CIRR, we leverage its unique design to report $\text{Recall}_{\text{subset}}@K$ ($R_{\text{subset}}@K$) with $K$ in $\{1, 2, 3\}$. This subset metric measures the ability to retrieve the correct item from a challenging, curated subset of six candidates, offering a more targeted measure of discriminative power.

\subsection{Implementation Details}
We use Qwen2.5-VL-7B~\cite{bai2025qwen2} as the backbone of ReCALL and fine-tune it with LoRA~\cite{hu2021lora} (rank $r{=}16$) on 8 NVIDIA H20 GPUs. Unless otherwise specified, we share the same training configuration across all stages. For FashionIQ, we use a learning rate of $4{\times}10^{-5}$, InfoNCE temperature of $\tau{=}0.03$, and a global batch size of 512, running 200 optimization steps for Stage~1 and 250 steps for Stage~4. For CIRR, we adopt a learning rate of $2{\times}10^{-5}$, $\tau{=}0.02$, and the same batch size, with 300 and 350 steps in Stage~1 and Stage~4 respectively. The triplet loss margin is $m{=}0.05$, and the weight $\lambda$ is $0.30$ on FashionIQ and $0.25$ on CIRR.

\subsection{Comparison with State-of-the-Art Methods} \label{sec:comparison}
We compare our proposed ReCALL framework against existing state-of-the-art methods on both CIRR and FashionIQ benchmarks, covering both traditional dual-tower approaches and recent MLLM-based retrievers.

\vspace{6pt}
\noindent\textbf{Results on CIRR.}
Table~\ref{tab:cirr} reports the quantitative results on the CIRR test set. The baseline ($\mathcal{R}_{\text{base}}$) alone delivers a competitive 51.23\% on $R@1$, confirming the inherent potential of MLLM architectures for compositional reasoning. Building on this, ReCALL establishes a new state-of-the-art of \textbf{55.52\%}, outperforming the concurrent MLLM-based CIR-LVLM~\cite{sun2025leveraging} (53.64\%). Notably, this relative improvement of \textbf{8.38\%} on $R@1$ over $\mathcal{R}_{\text{base}}$ compellingly validates the effectiveness of our \emph{diagnose--generate--refine} pipeline in rectifying capability degradation. Furthermore, on the $\text{Recall}_{\text{subset}}$ metrics designed for fine-grained evaluation, ReCALL secures a leading $R_{\text{subset}}@1$ of \textbf{81.49\%}. These gains confirm that our synthesized triplets successfully sharpen the model's decision boundaries against highly confounding visual distractors.

\vspace{6pt}
\noindent\textbf{Results on FashionIQ.}
Table~\ref{tab:fiq} details the quantitative results on the FashionIQ validation set. Despite inherent challenges such as high label noise and subtle attribute manipulations, ReCALL demonstrates consistent superiority by achieving the highest average $R@10$ of \textbf{57.04\%} and $R@50$ of \textbf{76.42\%}, successfully outperforming the concurrent CIR-LVLM~\cite{sun2025leveraging}. When compared to our $\mathcal{R}_{\text{base}}$, ReCALL delivers a robust \textbf{7.16\%} relative improvement in average $R@10$, with gains reaching as high as \textbf{10.71\%} in the \textit{Dress} category. These pervasive improvements across all categories compellingly validate that our minimal corrective editing strategy effectively captures nuanced visual-semantic distinctions, enabling precise retrieval even when target images differ from references by only fine-grained details.

\begin{table*}
  \centering
  \definecolor{dspgreen}{rgb}{0.0, 0.6, 0.0}
  \setlength{\tabcolsep}{3pt}
  \setlength{\aboverulesep}{0pt}
  \setlength{\belowrulesep}{0pt}
  \renewcommand{\arraystretch}{1.02} 
  
  \caption{Performance comparison on the FashionIQ validation set. We compare the proposed \textbf{ReCALL} ($\mathcal{R}_{\text{refine}}$) against state-of-the-art methods in terms of $\text{Recall}@K$ (\%). Consistent with Table~\ref{tab:cirr}, $\mathcal{R}_{\text{base}}$ denotes the baseline retriever obtained after \textbf{Stage~1}, serving as the starting point for recalibration. Best results are in \textbf{bold}, and the second-best are \underline{underlined}. The bottom row ($\Delta$) highlights the \textit{relative improvement} of ReCALL over $\mathcal{R}_{\text{base}}$.}
  \label{tab:fiq}
  \small
  \begin{tabular}{l c cc cc cc cc}
    \toprule
    \multirow{2}{*}{Method} & \multirow{2}{*}{Venue} & \multicolumn{2}{c}{Dress} & \multicolumn{2}{c}{Shirt} & \multicolumn{2}{c}{Top\&Tee} & \multicolumn{2}{c}{Avg.} \\
    \cmidrule(lr){3-4}\cmidrule(lr){5-6}\cmidrule(lr){7-8}\cmidrule(lr){9-10}
     & & $R@10$ & $R@50$ & $R@10$ & $R@50$ & $R@10$ & $R@50$ & $R@10$ & $R@50$ \\
    \midrule
    TIRG~\cite{Vo2019TIRG}       & \textcolor{gray}{CVPR'19} & 14.13 & 34.61 & 13.10 & 30.91 & 14.79 & 34.37 & 14.01 & 33.30 \\
    ARTEMIS~\cite{Delmas2022ARTEMIS} & \textcolor{gray}{ICLR'22} & 25.68 & 51.05 & 21.57 & 44.13 & 28.59 & 55.06 & 25.28 & 50.08 \\
    FashionSAP~\cite{Han2023FashionSAPSA} & \textcolor{gray}{CVPR'23} & 33.71 & 60.43 & 41.91 & 70.93 & 33.17 & 61.33 & 36.26 & 64.23 \\
    FAME-ViL~\cite{Han2023FAMEViLMV}   & \textcolor{gray}{CVPR'23} & 42.19 & 67.38 & 47.64 & 68.79 & 50.69 & 73.07 & 46.84 & 69.75 \\
    SyncMask~\cite{Song2024SyncMaskSA} & \textcolor{gray}{CVPR'24} & 33.76 & 61.23 & 35.82 & 62.12 & 44.82 & 72.06 & 38.13 & 65.14 \\
    SADN~\cite{wang2024semantic}       & \textcolor{gray}{MM'24}   & 40.01 & 65.10 & 43.67 & 66.05 & 48.04 & 70.93 & 43.91 & 67.36 \\
    CaLa~\cite{jiang2024cala}         & \textcolor{gray}{SIGIR'24} & 42.38 & 66.08 & 46.76 & 68.16 & 50.93 & 73.42 & 46.69 & 69.22 \\
    CoVR-2~\cite{ventura2024covr}     & \textcolor{gray}{TPAMI'24} & 46.53 & 69.60 & 51.23 & 70.64 & 52.14 & 73.27 & 49.96 & 71.17 \\
    SPRC~\cite{bai2023sentence}       & \textcolor{gray}{ICLR'24}  & 49.18 & 72.43 & 55.64 & 73.89 & 59.35 & 78.58 & 54.72 & 74.97 \\
    CIR-LVLM~\cite{sun2025leveraging} & \textcolor{gray}{AAAI'25}  & \underline{50.42} & \textbf{73.60} & \textbf{58.59} & \underline{75.86} & \underline{59.61} & \underline{78.99} & \underline{56.21} & \underline{76.14} \\
    CCIN~\cite{tian2025ccin}          & \textcolor{gray}{CVPR'25}  & 49.38 & 72.58 & 55.93 & 74.14 & 57.93 & 77.56 & 54.41 & 74.76 \\
    TME~\cite{li2025learning}         & \textcolor{gray}{CVPR'25}  & 49.73 & 71.69 & 56.43 & 74.44 & 59.31 & 78.94 & 55.15 & 75.02 \\
    QuRe~\cite{kwak2025qure}          & \textcolor{gray}{ICML'25}  & 46.80 & 69.81 & 53.53 & 72.87 & 57.47 & 77.77 & 52.60 & 73.48 \\
    \midrule
    Baseline ($\mathcal{R}_{\text{base}}$)          & - & 46.80 & 70.60 & 55.00 & 74.39 & 57.88 & 78.12 & 53.23 & 74.37 \\
    \rowcolor{gray!15}
    \textbf{ReCALL} ($\mathcal{R}_{\text{refine}}$) & - & \textbf{51.81} & \underline{73.48} & \underline{58.49} & \textbf{76.59} & \textbf{60.83} & \textbf{79.19} & \textbf{57.04} & \textbf{76.42} \\
    \midrule
    \multicolumn{2}{l}{\textit{Improvement ($\Delta$)}} 
    & \textit{\textcolor{dspgreen}{+10.71\%}} 
    & \textit{\textcolor{dspgreen}{+4.08\%}} 
    & \textit{\textcolor{dspgreen}{+6.35\%}} 
    & \textit{\textcolor{dspgreen}{+2.96\%}} 
    & \textit{\textcolor{dspgreen}{+5.10\%}} 
    & \textit{\textcolor{dspgreen}{+1.37\%}} 
    & \textit{\textcolor{dspgreen}{+7.16\%}} 
    & \textit{\textcolor{dspgreen}{+2.76\%}} \\
    \bottomrule
  \end{tabular}
\end{table*}

\begin{table}[t]
  \centering
  \small 
  \setlength{\tabcolsep}{6pt} 
  \setlength{\aboverulesep}{0pt}
  \setlength{\belowrulesep}{0pt}
  \renewcommand{\arraystretch}{1.1} 

  \caption{
    Ablation study on the mining strategy on the FashionIQ validation set. 
    We compare our \textbf{Self-Guided Mining} against a \textbf{Random Mining} baseline under the same data budget.
    To ensure statistical robustness, results for the Random strategy are averaged over \textbf{four independent runs} with different random seeds.
  }
  \label{tab:ablation_sampling}

  \resizebox{\linewidth}{!}{
  \begin{tabular}{l ccc}
    \toprule
    \textbf{Mining Strategy} & \textbf{R@10} & \textbf{R@50} & \textbf{Mean} \\
    \midrule
    
    $\mathcal{R}_{base}$ & 
    53.23\phantom{$_{\pm 0.00}$} & 
    74.37\phantom{$_{\pm 0.00}$} & 
    63.80\phantom{$_{\pm 0.00}$} \\
    
    + Random Mining & 
    53.80$_{\pm 0.20}$ & 
    74.32$_{\pm 0.06}$ & 
    64.06$_{\pm 0.10}$ \\
    
    \rowcolor{gray!15}
    + \textbf{Self-Guided} & 
    \textbf{57.04}\phantom{$_{\pm 0.00}$} & 
    \textbf{76.42}\phantom{$_{\pm 0.00}$} & 
    \textbf{66.73}\phantom{$_{\pm 0.00}$} \\
    \bottomrule
  \end{tabular}
  }
\end{table}
\begin{table}[t]
  \centering
  \small 
  \setlength{\tabcolsep}{6pt} 
  \definecolor{dspgreen}{rgb}{0.0, 0.6, 0.0} 
  \setlength{\aboverulesep}{0pt}
  \setlength{\belowrulesep}{0pt}
  \renewcommand{\arraystretch}{1.1} 
  
  \caption{
    Ablation study of the core components on the FashionIQ validation set. 
    \textbf{CG}: CoT-assisted Generation, 
    \textbf{VC}: VQA-Assisted Quality Control, 
    \textbf{GR}: Grouped Contrastive Refinement. 
    $\bullet$ denotes the component is included, and $\circ$ denotes excluded. 
    All metrics are the average over the three categories (in \%).
    The stepwise performance improvements validate the effectiveness of each proposed module.
  }
  \label{tab:ablation_components}
  
  \begin{tabular}{c ccc ccc}
    \toprule
    \multirow{2}{*}{\textbf{Baseline}} & \multicolumn{3}{c}{\textbf{Components}} & \multicolumn{3}{c}{\textbf{Metrics (Avg.)}} \\
    \cmidrule(lr){2-4} \cmidrule(lr){5-7}
    
     & \textbf{CG} & \textbf{VC} & \textbf{GR} & \textbf{R@10} & \textbf{R@50} & \textbf{Mean} \\
    \midrule
    
    $\bullet$ & $\circ$ & $\circ$ & $\circ$ & 53.23 & 74.37 & 63.80 \\
    
    $\bullet$ & $\bullet$ & $\circ$ & $\circ$ & 55.41 & 75.17 & 65.29 \\
    
    $\bullet$ & $\bullet$ & $\bullet$ & $\circ$ & 56.13 & 76.04 & 66.09 \\
    
    \rowcolor{gray!15}
    $\bullet$ & $\bullet$ & $\bullet$ & $\bullet$ & \textbf{57.04} & \textbf{76.42} & \textbf{66.73} \\
    
    \bottomrule
  \end{tabular}
\end{table}

\begin{figure}[t]
  \centering
  \vspace{-5pt}
  \includegraphics[width=0.8\linewidth]{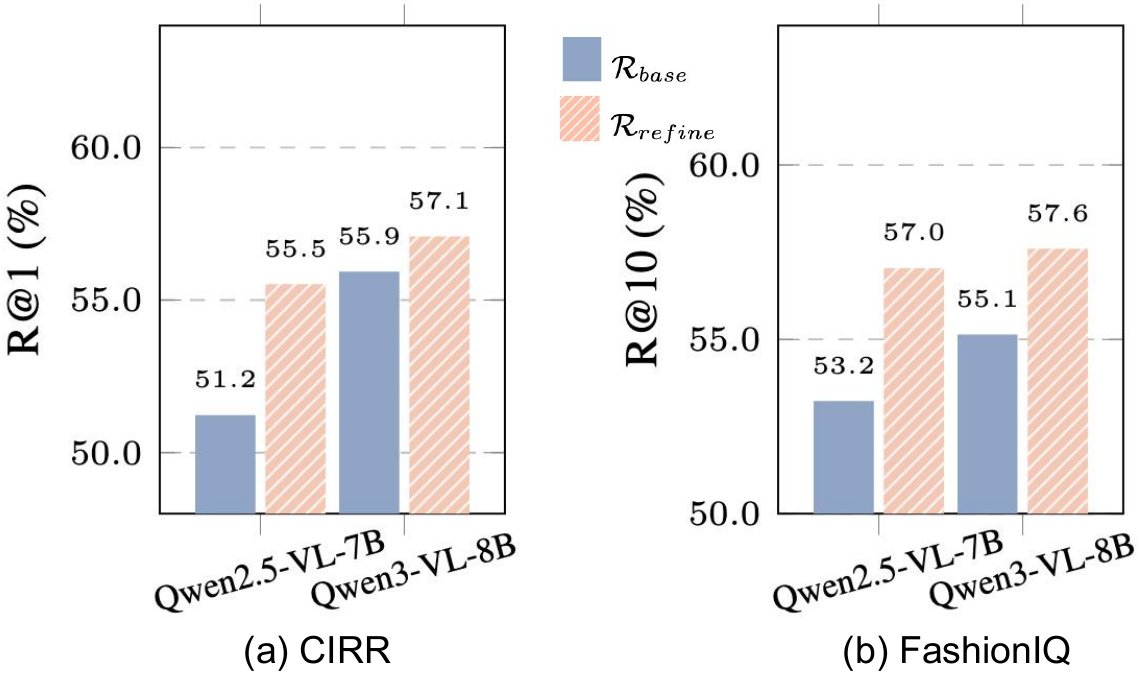}
  \caption{
  \textbf{Generalizability across backbones.} 
  We validate ReCALL on different foundation models (\textbf{Qwen2.5-VL-7B} and \textbf{Qwen3-VL-8B}). 
  Despite higher baselines, ReCALL consistently delivers performance gains on both (a) CIRR and (b) FashionIQ, confirming the strong generalizability of our framework.}
  \label{fig:model_agnostic}
\end{figure}

\subsection{Ablation Studies} 
\label{sec:ablation}
We conduct a series of experiments to validate the efficacy, efficiency, and generalization capabilities of ReCALL.
\vspace{4pt}

\noindent\textbf{Diagnose Phase: Impact of Self-Guided Informative Instance Mining.}
We further investigate the necessity of the Diagnose phase. A prevailing trend in recent MLLM adaptation involves indiscriminate large-scale data synthesis. To strictly simulate this scaling approach, we establish a \emph{Random Mining} baseline. Specifically, for every training query, we first retrieve the top-50 candidate images using the frozen $\mathcal{R}_{\text{base}}$. From this candidate pool, we randomly sample negative instances to undergo the generation pipeline, strictly maintaining the same data scale as our method. To guarantee experimental robustness, we report the mean and standard deviation across four independent runs (using different random seeds) in Table~\ref{tab:ablation_sampling}. The results reveal a critical inefficiency in the blind synthesis paradigm. Even when averaged over multiple runs, Random Mining yields only marginal gains (improving $R@10$ from 53.23\% to 53.80\%), whereas our Self-Guided strategy delivers a substantial boost to 57.04\%. This remarkable contrast demonstrates that indiscriminate synthesis often results in severe redundancy: since the candidates are drawn randomly from the top-50, many have likely already been correctly ranked by the model, thus providing negligible gradient signals. In contrast, ReCALL follows a \emph{diagnose--then--generate} philosophy, precisely concentrating the generative budget on the model's active failure cases. By ensuring that every synthesized triplet targets a specific cognitive deficiency, ReCALL achieves superior capability enhancement with maximal data efficiency.

\begin{figure*}[t]
  \centering
  \includegraphics[width=0.72\textwidth]{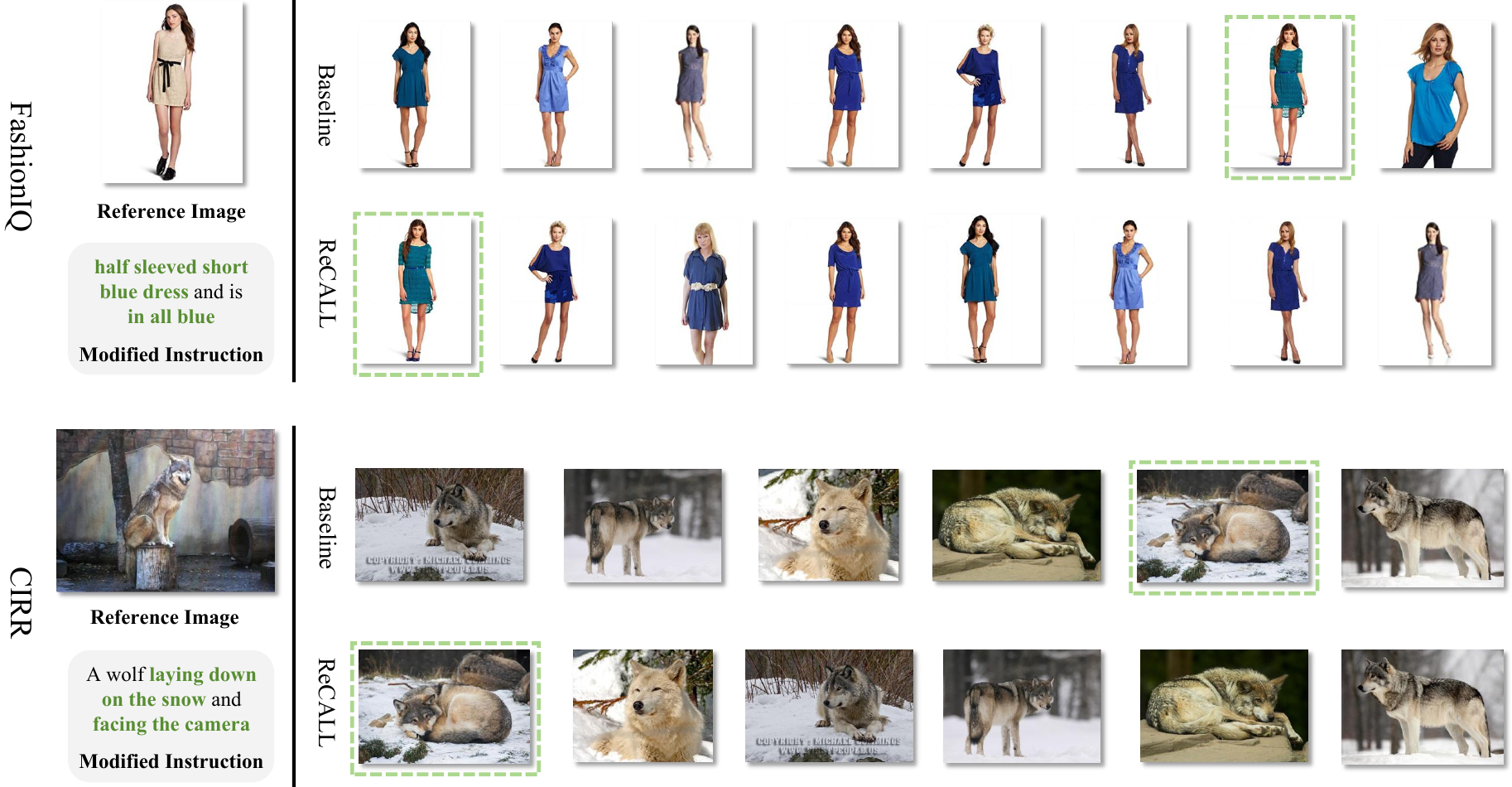}
  \caption{Qualitative comparison between the baseline ($\mathcal{R}_{\text{base}}$) and our ReCALL ($\mathcal{R}_{\text{refine}}$) on FashionIQ (top) and CIRR (bottom). The green dashed boxes indicate the ground-truth targets. $\mathcal{R}_{\text{base}}$ suffers from capability degradation, failing to capture specific details like ``half sleeved'' or ``facing the camera,'' while ReCALL successfully retrieves the correct targets by recalibrating the fine-grained reasoning.}
  \label{fig:qualitative}
\end{figure*}

\vspace{2pt}
\noindent\textbf{Generate Phase: Effectiveness of Generative Calibration (CG \& VC).}
This set of experiments verifies the core \emph{Generate} phase using a progressive study detailed in Table~\ref{tab:ablation_components} (Rows 1-3). We leverage the foundation model $\mathcal{F}$ to create corrective supervision. The introduction of CoT-assisted Generation (CG) yields a substantial gain, boosting $R@10$ from 53.23\% to 55.41\%. This absolute improvement of 2.18\% confirms that capitalizing on the native generative reasoning of $\mathcal{F}$ to synthesize targeted supervision effectively mitigates cognitive deficits. Furthermore, adding VQA-Assisted Quality Control (VC) further elevates $R@10$ to 56.13\%. This step utilizes the intrinsic discriminative understanding of $\mathcal{F}$ to filter out noise, ensuring that only high-quality triplets guide the training. Overall, these results empirically demonstrate that our framework successfully internalizes the robust compositional reasoning abilities of the foundation model, alleviating capability degradation caused by the initial adaptation.

\vspace{2pt}
\noindent\textbf{Refine Phase: Necessity of Grouped Refinement (GR).}
We finally validate the \emph{Refine} phase (Table~\ref{tab:ablation_components}, Rows 3--4), which focuses on how to effectively internalize the corrective supervision. As shown in Row 3, merely expanding the training set with synthetic triplets via standard random batching offers limited gains. By contrast, enabling Grouped Contrastive Refinement (GR) achieves the peak performance of 57.04\% on $R@10$. This comparison highlights the importance of optimal data utilization: our grouped strategy is specifically designed to leverage the subtle visual and textual contrasts created in the generation phase. By forcing a direct, in-batch comparison between the target and its synthesized near-neighbor, this mechanism compels the model to explicitly resolve ambiguities within the micro-group. This optimal signal transmission effectively translates the corrective supervision into sharper, fine-grained discriminative boundaries, successfully recalibrating the degraded compositional reasoning capability.

\vspace{2pt}
\noindent\textbf{Generalizability across Backbones.}
To verify that ReCALL is a model-agnostic framework rather than a specific patch for weaker architectures, we applied our method to a more advanced foundation model, \textbf{Qwen3-VL-8B}. As illustrated in Fig.~\ref{fig:model_agnostic}, the baseline adaptation ($\mathcal{R}_{\text{base}}$) of Qwen3-VL-8B already exhibits a strong starting point, significantly outperforming the standard Qwen2.5-VL-7B baseline (e.g., 55.93\% vs. 51.23\% on $R@1$ for CIRR). Despite this high baseline, applying ReCALL still yields consistent improvements, boosting $R@1$ on CIRR to 57.09\% and $R@10$ on FashionIQ to 57.60\%. It is worth noting that even as the foundation model becomes stronger, the \emph{capability degradation} phenomenon stemming from the paradigm conflict between generation and retrieval persists. Our results confirm that ReCALL effectively addresses this fundamental issue, demonstrating scalability and robustness across different model capacities.

\subsection{Qualitative Analysis}
\label{sec:qualitative}

To visually demonstrate the \emph{capability degradation} and the effectiveness of our subsequent recalibration, Fig.~\ref{fig:qualitative} presents two representative challenging cases from the FashionIQ and CIRR datasets that demand precise fine-grained reasoning. Crucially, we verify that the foundation model ($\mathcal{F}$) correctly identifies both targets via VQA reasoning, confirming that the requisite compositional knowledge already exists in the pre-trained model. However, the adapted $\mathcal{R}_{base}$ fails in both instances, exposing a clear degradation pattern. The baseline retains coarse-grained understanding, such as identifying a ``blue dress'' or a ``wolf on snow,'' but collapses on specific constraints. For example, it retrieves a sleeveless dress instead of the requested ``half sleeved'' one, and a profile-view wolf ignoring the instruction ``facing the camera.'' In contrast, ReCALL successfully rectifies these errors. By diagnosing these blind spots and realigning the representation space with the native reasoning ability of the foundation model, our method restores the lost sensitivity to subtle attributes and spatial relations, accurately retrieving the correct targets in both scenarios.

\section{Conclusion}
\label{sec:conclusion}

In this work, we address \emph{capability degradation}---the deterioration of fine-grained reasoning when adapting generative MLLMs for retrieval---by proposing ReCALL. Our framework utilizes the intrinsic zero-shot reasoning of MLLMs via a \emph{diagnose--generate--refine} pipeline to create and internalize targeted corrective supervision. Specifically, self-guided informative instance mining and grouped refinement embed the foundation model's reasoning into the retrieval space. Empirical results demonstrate that ReCALL achieves state-of-the-art performance on mainstream CIR benchmarks.

\section*{Acknowledgments}
This work was supported in part by the National Key R\&D Program of China (No. 2022ZD0160601), the National Natural Science Foundation of China under Grants 62276260 and U1701266, the Beijing Natural Science Foundation (Grant No. L252035), and the Guangdong Provincial Key Laboratory of Intellectual Property and Big Data under Grant 2018B030322016.

{
    \small
    \bibliographystyle{ieeenat_fullname}
    \bibliography{main}
}

\clearpage
\setcounter{page}{1}
\maketitlesupplementary

\section{Additional Experimental Results and Analysis}
\label{sec:appendix_exp}

\subsection{Data Scale Study on FashionIQ}
\label{sec:appendix_scale}

To investigate the scalability of our Self-Guided Informative Instance Mining strategy, we conduct a quantitative analysis on the FashionIQ dataset by varying the mining hyperparameter $K$ (denoted as top-$K$). This parameter determines the maximum number of informative instances mined for each failure query, directly controlling the volume of synthesized supervision.

\noindent\textbf{Experimental Setup.} To decouple data scaling from quality filtering, we conduct experiments \textit{without} the VQA-Assisted Quality Control mechanism. The model is trained via the standard InfoNCE loss within our Grouped Contrastive Refinement framework.

\begin{figure}[h]
  \centering
  \includegraphics[width=\linewidth]{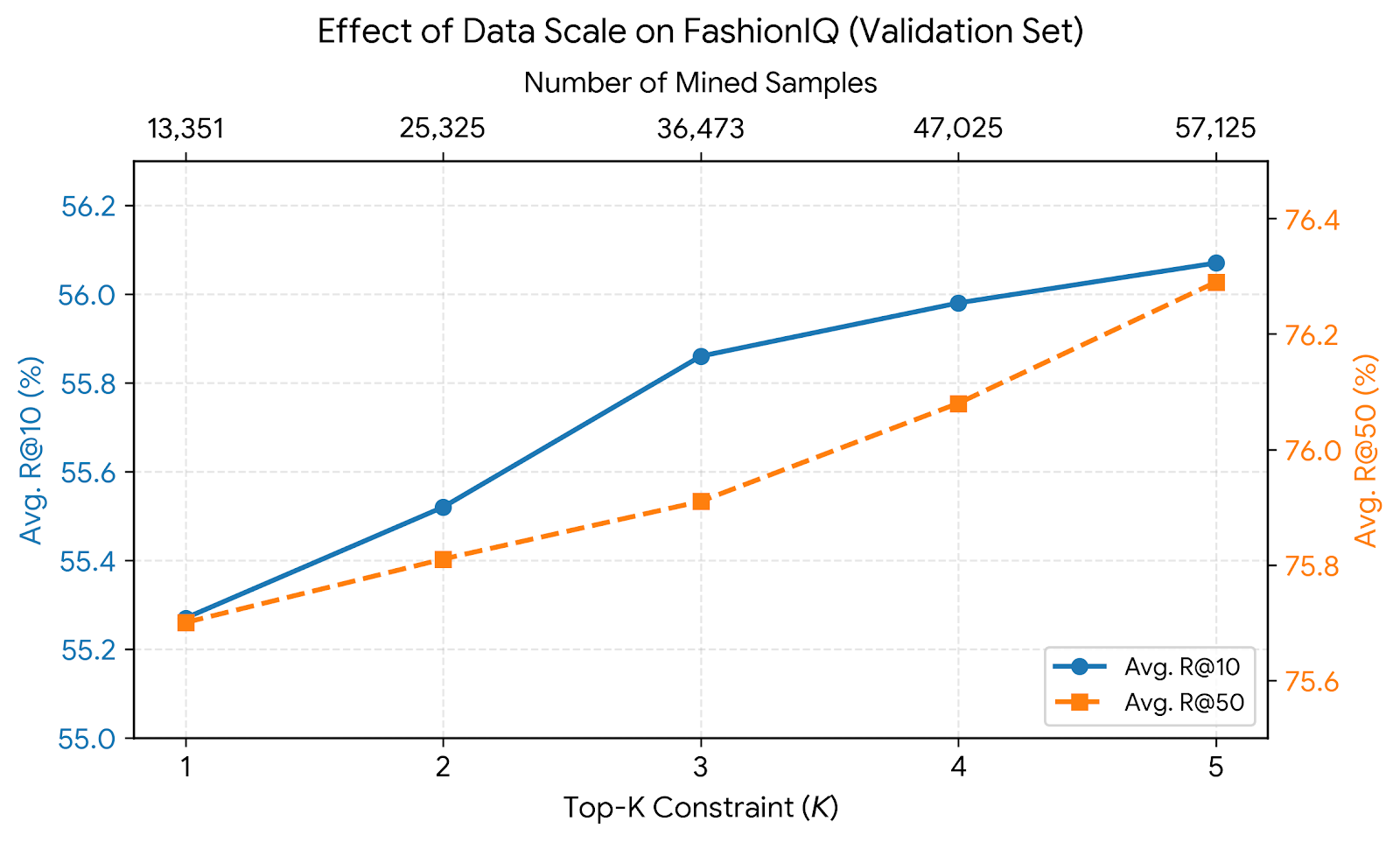}
  \caption{\textbf{Effect of data scale on the FashionIQ validation set.} The visualization employs dual x-axes to map the mining hyperparameter $K$ (bottom) to the corresponding number of mined samples (top). The dual y-axes (left for R@10, right for R@50) with zoomed-in scales highlight the monotonic performance gains as the data scale increases.}
  \label{fig:data_scale}
\end{figure}

\noindent\textbf{Results.} The quantitative results are visualized in \cref{fig:data_scale}. We employ a dual-axis plot to illustrate the relationship between the mining constraint $K$ (bottom x-axis) and the resultant volume of synthesized training samples (top x-axis).
As illustrated, increasing $K$ from 1 to 5 significantly expands the training set from 13,351 to 57,125 samples. Crucially, this increase in data scale correlates with a consistent upward trend in retrieval performance. Specifically, Avg. R@10 (left axis) improves from 55.27\% to 56.07\%, and Avg. R@50 (right axis) rises from 75.70\% to 76.29\%. This positive scaling effect demonstrates that ReCALL can effectively leverage larger pools of informative instances to refine its discriminative boundaries, yielding continuous gains even in the absence of additional filtering.

\subsection{Hyperparameter Analysis of Triplet Loss}
\label{sec:appendix_triplet}

To identify the optimal configuration for the targeted refinement stage, we conduct a grid search over two critical hyperparameters in the joint loss function: the triplet loss weight $\lambda$ and the margin $m$. We evaluate the model on the FashionIQ validation set, identifying the optimal setting based on the Average R@10 metric. Specifically, the weight $\lambda$ is varied within $\{0.1, 0.2, 0.3, 0.4, 0.5\}$, and the margin $m$ within $\{0.05, 0.10, 0.20\}$.

\begin{figure*}[t]
  \centering
  \includegraphics[width=0.85\textwidth]{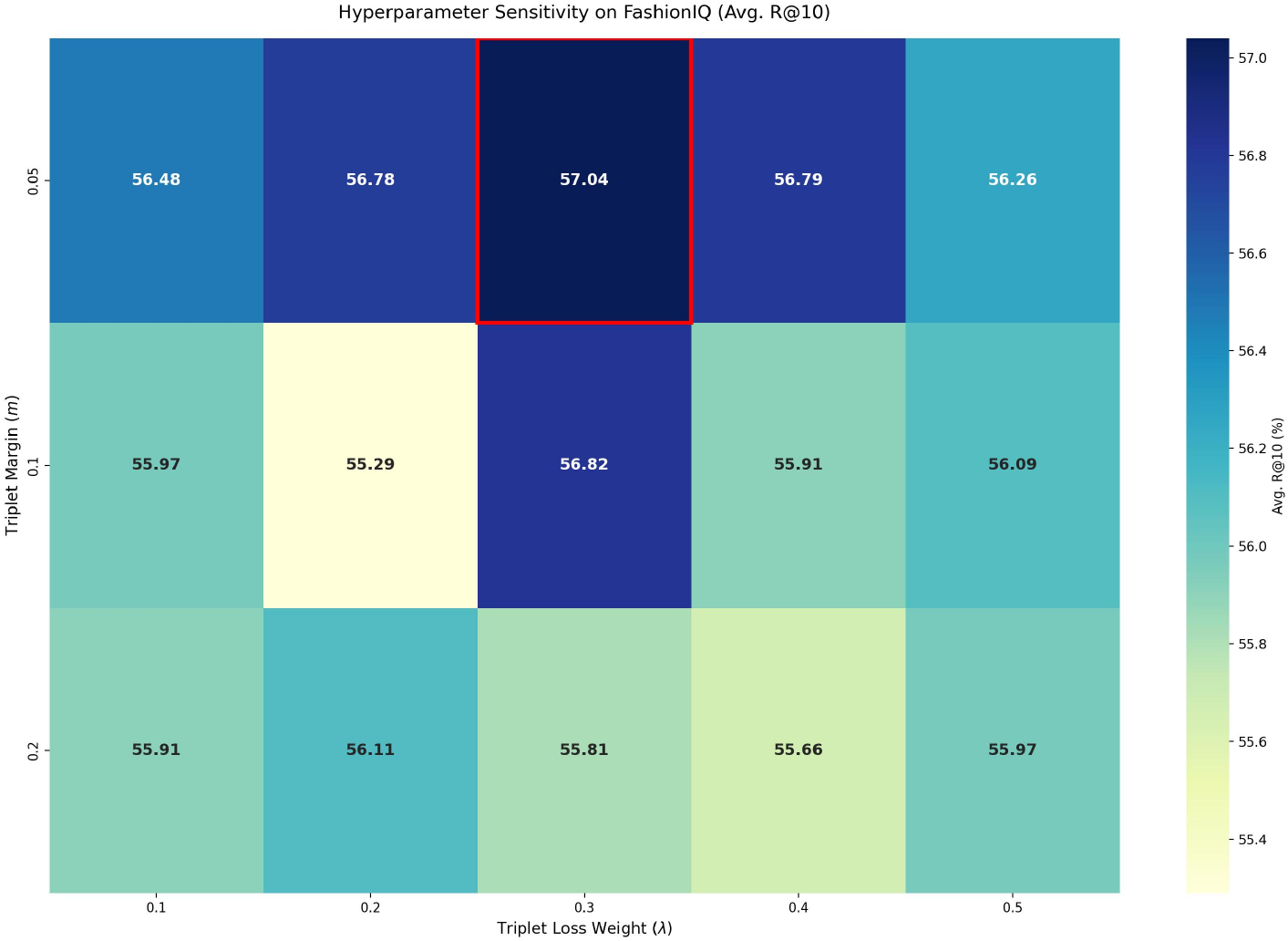}
  \caption{\textbf{Hyperparameter sensitivity analysis on the FashionIQ validation set.} We report Avg. R@10 (\%) under varying triplet loss weights ($\lambda$) and margins ($m$). The red box highlights the optimal configuration adopted in our final model.}
  \label{fig:triplet_heatmap}
\end{figure*}

\noindent\textbf{Results.} The sensitivity analysis is visualized in \cref{fig:triplet_heatmap}.
First, concerning the loss weight $\lambda$, performance generally peaks at $\lambda=0.3$. Lower weights (e.g., $\lambda=0.1$) provide insufficient supervision for fine-grained discrimination, whereas excessive weights (e.g., $\lambda=0.5$) tend to over-regularize the representation, potentially conflicting with the global alignment objective of the InfoNCE loss.
Second, regarding the margin $m$, the model consistently favors a tighter constraint ($m=0.05$). This preference suggests that the informative instances mined by our framework share high visual affinity with the ground truth targets. Consequently, a tighter margin compels the model to resolve these fine-grained ambiguities without disrupting the broader semantic structure of the embedding space.
Based on these empirical findings, we adopt $\lambda=0.3$ and $m=0.05$ as the default configuration for FashionIQ, which yields the best Avg. R@10 of 57.04\%.

\subsection{Computational Cost and Efficiency Analysis}
\label{sec:appendix_cost}

To ensure reproducibility and transparency regarding resource utilization, we detail the computational costs and data statistics \textbf{of} the ReCALL framework. All experiments were conducted on 8 NVIDIA H20 GPUs. \cref{tab:time_cost} summarizes the training duration, generation latency, and filtering statistics for both the CIRR and FashionIQ datasets.

\vspace{5pt}
\noindent\textbf{Analysis.} We analyze the computational overhead across the three primary phases of our pipeline:

\noindent\textbf{Comparable Training Latency (Stage 1 vs. Stage 4).} The training duration for Targeted Refinement (Stage 4) is virtually identical to that of the Baseline Adaptation (Stage 1). For instance, on CIRR, Stage 4 requires approximately 3.6 hours, matching the 3.6 hours of Stage 1. This equivalence demonstrates that our Grouped Contrastive Refinement strategy (Sec.~\ref{sec:stage4}) effectively recalibrates the model without introducing significant computational overhead to the online training loop.

\vspace{2pt}
\noindent\textbf{One-off Offline Synthesis (Stages 2 \& 3).} The combined process of mining informative instances and synthesizing corrective instructions constitutes the primary computational cost. Specifically, CoT-assisted generation accounts for approximately 14.2 hours on CIRR and 10.9 hours on FashionIQ. Crucially, however, this represents a \textit{one-time, offline investment}. Once synthesized, these high-quality triplets serve as a permanent asset that can be reused indefinitely for subsequent training runs or hyperparameter tuning, rendering the amortized cost negligible.

\vspace{2pt}
\noindent\textbf{Efficient Quality Assurance.} The VQA-Assisted Quality Control mechanism demonstrates high efficiency. It effectively purges noisy data—removing 5,455 instances (8.5\%) for CIRR and 5,947 instances (10.4\%) for FashionIQ—while consuming only $\sim$1 hour of processing time. This ensures that the final refinement is driven by high-fidelity supervision with minimal time penalty.

\begin{table*}[t]
  \centering
  \caption{\textbf{Detailed statistics of computational cost and data generation across the ReCALL pipeline.} Generation denotes the CoT-assisted synthesis in Stage 3, while Filtering refers to the VQA-based quality control. Note that the costs associated with Stages 2 and 3 are \textbf{one-time and offline}.}
  \label{tab:time_cost}
  \resizebox{\linewidth}{!}{
  \begin{tabular}{l|c|ccc|c}
    \toprule
    \multirow{2}{*}{\textbf{Dataset}} & \textbf{Stage 1} & \multicolumn{3}{c|}{\textbf{Stages 2 \& 3: Diagnose \& Generate (Offline)}} & \textbf{Stage 4} \\
    \cmidrule(lr){2-2} \cmidrule(lr){3-5} \cmidrule(lr){6-6}
     & \textit{Adaptation Time} & \textit{Generation Time} & \textit{Samples (Generated $\to$ \textbf{Kept})} & \textit{Filtering Time} & \textit{Refinement Time} \\
    \midrule
    \textbf{CIRR} & 3h 34m & $\sim$14.2h & 64,105 $\to$ \textbf{58,650} & 1h 13m & 3h 35m \\
    \textbf{FashionIQ} & 2h 43m & $\sim$10.9h & 57,125 $\to$ \textbf{51,155} & 1h 04m & 2h 45m \\
    \bottomrule
  \end{tabular}
  }
\end{table*}

\subsection{Generalization and Transferability Across Backbones}
\label{sec:appendix_transfer}

To evaluate the generalizability and transferability of the ReCALL framework, we extend our experiments to a different model family, LLaVA-NeXT. Furthermore, we investigate whether the informative instances and corrective instructions synthesized by one model can benefit another. 

As shown in \cref{tab:transferability}, cross-model transfer (training LLaVA-NeXT using triplets synthesized by Qwen2.5-VL) yields a +1.15\% gain (51.93\% $\rightarrow$ 53.08\%) on CIRR. This confirms that different MLLMs share certain common cognitive blind spots, allowing synthesized corrective data to be highly transferable. However, the full ReCALL pipeline—where LLaVA-NeXT diagnoses and refines its own specific failure cases—achieves a more significant gain of +2.72\% (54.65\% on R@1). This indicates that while cross-model transfer is effective, the self-diagnosis phase remains essential to optimally address model-specific cognitive gaps.

\begin{table}[h]
    \centering
    \caption{\textbf{Generalization \& Transferability Analysis on the CIRR test set.}}
    \label{tab:transferability}
    \resizebox{\linewidth}{!}{
    \begin{tabular}{@{}l cccc@{}}
        \toprule
        \textbf{Method Setting} & \textbf{R@1} & \textbf{R@5} & \textbf{R@10} & \textbf{R@50} \\
        \midrule
        LLaVA Baseline ($\mathcal{R}_{base}$) & 51.93 & 81.87 & 88.95 & 97.58 \\
        + ReCALL (Transfer: Qwen Data) & 53.08 & 83.40 & 91.21 & 98.46 \\
        + ReCALL (Full Pipeline) & \textbf{54.65} & \textbf{84.02} & \textbf{91.33} & \textbf{98.41} \\
        \bottomrule
    \end{tabular}
    }
\end{table}

\subsection{Comparison with Alternative Mining Strategies}
\label{sec:appendix_mining}

We further compare our Self-Guided Informative Instance Mining against a standard Hard Negative Mining baseline. For the Hard Negative baseline, we re-finetune $\mathcal{R}_{base}$ directly using the mined informative instances as hard negatives without any textual refinement.

As reported in \cref{tab:hmmbaseline}, the Hard Negative strategy achieves an R@1 of 52.57\%, comparable to a Random Mining strategy (52.07\%), and shows a notable decline in broader metrics (R@5/10/50) compared to the baseline. This implies that blindly enforcing repulsion on visually ambiguous negatives—without explicitly defining \textit{why} they differ—introduces contradictory gradients that distort the learned manifold. In contrast, ReCALL resolves this via \textit{semantic correction}: we generate $\tilde{T}_m$ to explicitly describe the hard negative, converting it into a constructive positive pair ($I_r, \tilde{T}_m, I_h$). This precise semantic direction explains the superior capability calibration (+4.29\% on R@1) achieved by the full ReCALL pipeline.

\begin{table}[h]
    \centering
    \caption{\textbf{Comparison of Mining Strategies on CIRR.}}
    \label{tab:hmmbaseline}
    \resizebox{\linewidth}{!}{
    \begin{tabular}{@{}l cccc@{}}
        \toprule
        \textbf{Method Setting} & \textbf{R@1} & \textbf{R@5} & \textbf{R@10} & \textbf{R@50} \\
        \midrule
        Baseline ($\mathcal{R}_{base}$) & 51.23 & 82.15 & 90.20 & 98.20 \\
        + Random Mining & 52.07 & 81.64 & 90.02 & 97.84 \\
        + Hard Neg. Mining (No Edit) & 52.57 & 81.56 & 89.33 & 97.70 \\
        + ReCALL (Full Pipeline) & \textbf{55.52} & \textbf{84.07} & \textbf{91.83} & \textbf{98.55} \\
        \bottomrule
    \end{tabular}
    }
\end{table}

\subsection{Ablation on Model Capacity and Adaptation}
\label{sec:appendix_lora}

To investigate whether capability degradation stems from limited parameter capacity during adaptation, we conducted ablations by scaling the LoRA rank ($r=32, 64$) and performing Full Fine-tuning. 

As shown in \cref{tab:lora_ablation}, increasing the number of trainable parameters paradoxically worsens retrieval performance. This confirms that capability degradation is not a consequence of limited parameter capacity. Instead, it originates from the intrinsic paradigm conflict between the MLLM's generative pre-training and the discriminative retrieval adaptation. Under a fixed training dataset, expanding trainable parameters accelerates overfitting to the coarse-grained retrieval task, thereby exacerbating the suppression of native fine-grained reasoning priors.

\begin{table}[h]
    \centering
    \caption{\textbf{Ablation on LoRA Rank \& Full Fine-tuning on CIRR.}}
    \label{tab:lora_ablation}
    \resizebox{\linewidth}{!}{
    \begin{tabular}{@{}l cccc@{}}
        \toprule
        \textbf{Setting} & \textbf{R@1} & \textbf{R@5} & \textbf{R@10} & \textbf{R@50} \\
        \midrule
        LoRA $r=16$ (Ours Baseline) & \textbf{51.23} & \textbf{82.15} & \textbf{90.20} & \textbf{98.20} \\
        LoRA $r=32$ & 51.04 & 81.69 & 89.54 & 98.22 \\
        LoRA $r=64$ & 49.74 & 80.58 & 89.35 & 98.05 \\
        Full Fine-tuning & 48.70 & 80.55 & 89.64 & 97.98 \\
        \bottomrule
    \end{tabular}
    }
\end{table}

\subsection{Further Methodological Discussions}
\label{sec:appendix_discussion}

\noindent\textbf{Distribution Integrity and Label Space.} It is crucial to emphasize that ReCALL operates strictly as an informative instance augmentation strategy rather than altering or re-labeling the original ground-truth targets. The original triplets $(I_r, T_m, I_t)$ are strictly retained to anchor the model to the source distribution. Furthermore, our Minimal Edit Principle (Sec.~\ref{sec:stage3}) guarantees that the synthesized text $\tilde{T}_m$ matches the original style and length. This design ensures that ReCALL provides additive regularization to sharpen decision boundaries without shifting the training label space.

\vspace{2pt}
\noindent\textbf{Reliability of VQA-Assisted Filtering.} To quantitatively ensure the reliability of the generated corrective supervision, we conducted a rigorous human evaluation of the VQA-Assisted Quality Control mechanism (Sec.~\ref{sec:stage3}). We employed three human evaluators to verify 300 randomly sampled triplets that passed the VQA filter (confidence threshold $\geq$ 0.95). The evaluation yielded a high average accuracy of 92\%, confirming that the VQA-based check serves as a highly reliable proxy for filtering valid textual modifications.

\section{Prompt Details}
\label{sec:appendix_prompts}

\begin{figure*}[h]
  \centering
  \begin{subfigure}{0.48\linewidth}
    \includegraphics[width=\linewidth]{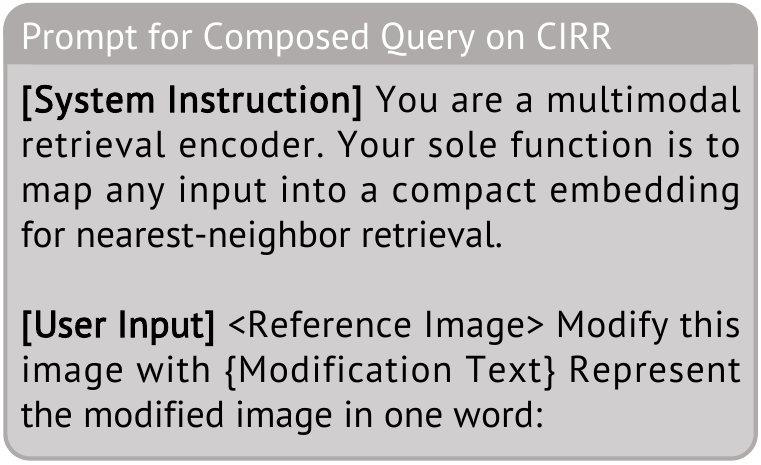}
    \caption{Prompt for Composed Query on CIRR}
    \label{fig:prompt_cirr_query}
  \end{subfigure}
  \hfill
  \begin{subfigure}{0.48\linewidth}
    \includegraphics[width=\linewidth]{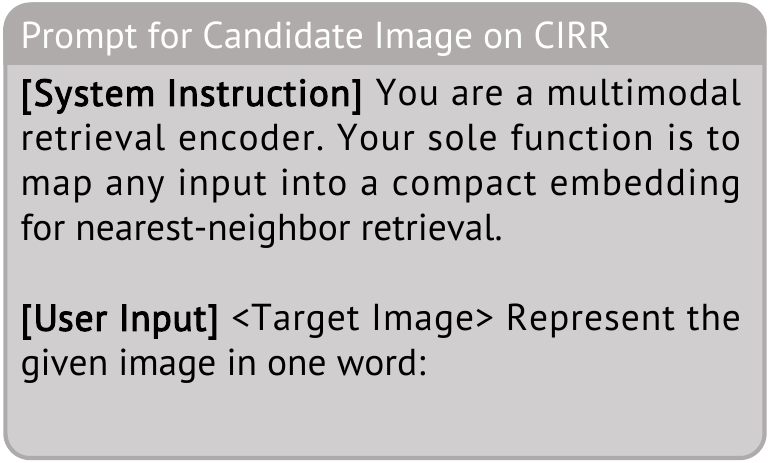}
    \caption{Prompt for Candidate Image on CIRR}
    \label{fig:prompt_cirr_candidate}
  \end{subfigure}
  \\ 
  \begin{subfigure}{0.48\linewidth}
    \includegraphics[width=\linewidth]{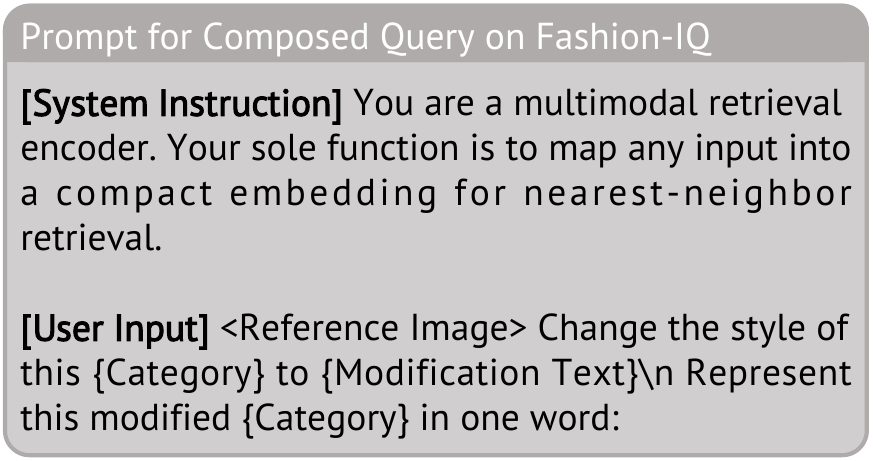}
    \caption{Prompt for Composed Query on FashionIQ}
    \label{fig:prompt_fiq_query}
  \end{subfigure}
  \hfill
  \begin{subfigure}{0.48\linewidth}
    \includegraphics[width=\linewidth]{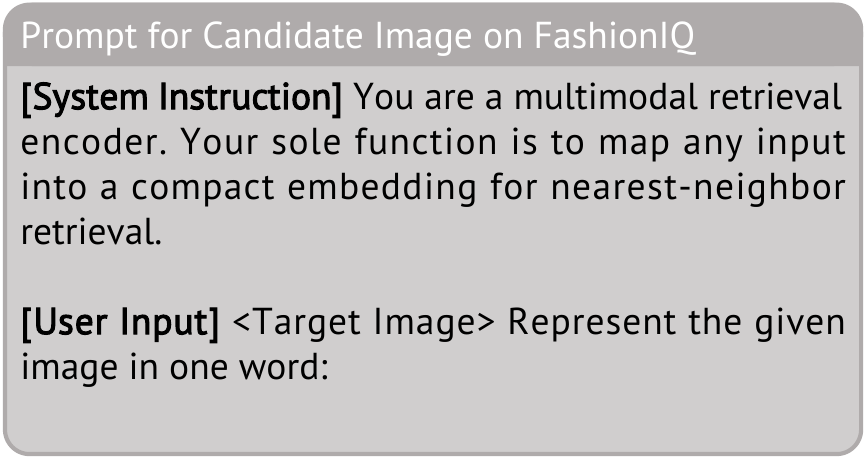}
    \caption{Prompt for Candidate Image on FashionIQ}
    \label{fig:prompt_fiq_candidate}
  \end{subfigure}
  
  \caption{Full prompt templates for retrieval encoding on CIRR and FashionIQ. The structure utilizes an integrated System Instruction to enforce the role of a discriminative encoder. The query prompts are specialized: CIRR uses a general modification instruction, while FashionIQ incorporates category information for fine-grained attribute manipulation.}
  \label{fig:retrieval_prompts}
\end{figure*}

\begin{figure}[h]
  \centering
  \includegraphics[width=\linewidth]{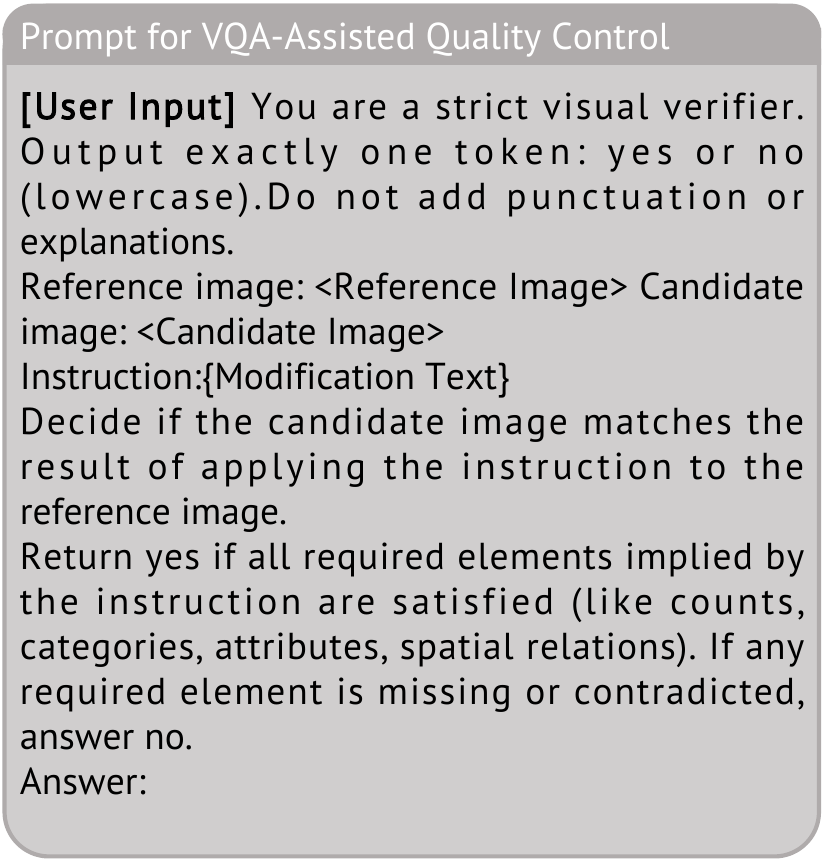}
  \caption{Prompt template for the VQA-Assisted Quality Control mechanism. This zero-shot prompt conditions the Foundation Model to act as a strict binary verifier, ensuring the quality of the synthesized informative triplets.}
  \label{fig:vqa_prompt}
\end{figure}

\subsection{Retrieval Prompts for Query and Candidate Encoding}
\label{sec:appendix_prompt_query}

To counteract the Capability Degradation identified in \cref{sec:intro}, we engineer specialized prompt templates that explicitly condition the MLLM to operate as a discriminative retrieval encoder ($\mathcal{R}_{base}$ and $\mathcal{R}_{refine}$), effectively suppressing its default conversational tendencies.

\cref{fig:retrieval_prompts} illustrates the prompt architectures employed for encoding inputs on both the CIRR and FashionIQ datasets. Our design adheres to two governing principles:

\noindent\textbf{Role Enforcement via System Instruction.} A mandatory system instruction is embedded in every prompt instance. This directive explicitly constrains the model's output space, enforcing a retrieval-oriented role and inhibiting open-ended generative behaviors.

\noindent\textbf{Dataset-Specific Attention Guidance.} The user input instruction is tailored to steer the model's attention mechanism towards feature fusion strategies appropriate for each dataset. We highlight a critical distinction in the Composed Query prompt: whereas the CIRR template employs a generalized modification instruction suitable for open-domain objects, the FashionIQ template integrates category-aware phrasing (e.g., \textit{``Change the style of this \{Category\}...''}) to enhance domain specificity and attribute sensitivity.

\subsection{Prompts for VQA-Assisted Quality Control}
\label{sec:appendix_prompt_vqa}

The generative calibration process described in \cref{sec:stage3} entails an inherent risk of synthesizing hallucinated or visually ungrounded corrective triplets. To attenuate this noise, we implement a VQA-Assisted Quality Control mechanism, repurposing the Foundation Model ($\mathcal{F}$) to function as a rigorous visual verifier. This step necessitates a specialized VQA prompt designed to validate the semantic alignment between the synthesized modified instruction ($\tilde{T}_m$) and the actual informative instance ($I_h$).

\cref{fig:vqa_prompt} illustrates the prompt structure engineered for this verification task. Our design relies on two key mechanisms:

\noindent\textbf{Strict Binary Constraint.} The prompt explicitly constrains the model's output space, mandating a single, lowercase token response (\texttt{yes} or \texttt{no}). This binary restriction inhibits the model's open-ended generative tendencies.

\noindent\textbf{Discriminative Reasoning Activation.} By disabling the generative mode, the constraint compels the model to perform critical discriminative reasoning to verify semantic consistency. This serves as a robust filter, ensuring that only high-fidelity informative instances are admitted into the final refinement stage.

\subsection{Prompts for CoT-Assisted Instruction Synthesis}
\label{sec:appendix_prompt_cot}

\begin{figure*}[p]
  \centering
  \subfloat[Prompt for CoT-Assisted Generation on CIRR]{\includegraphics[width=0.48\linewidth]{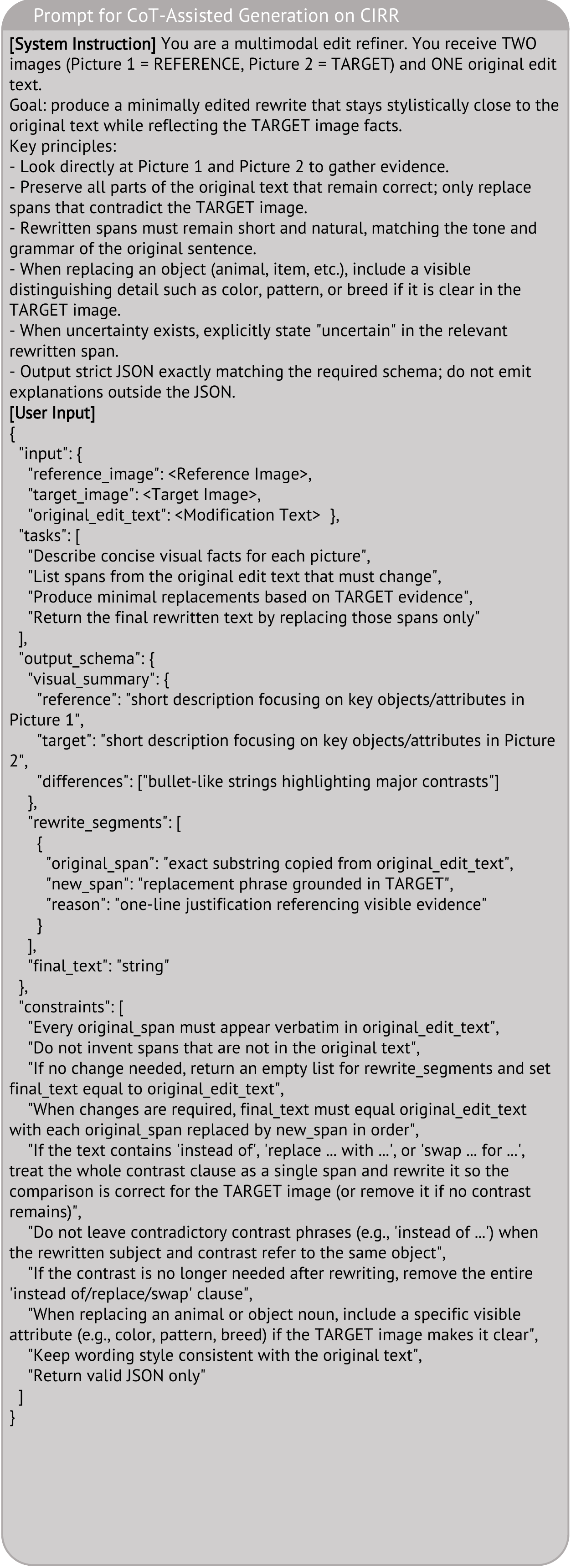}\label{fig:prompt_cot_cirr}} \hfill
  \subfloat[Prompt for CoT-Assisted Generation on FashionIQ]{\includegraphics[width=0.48\linewidth]{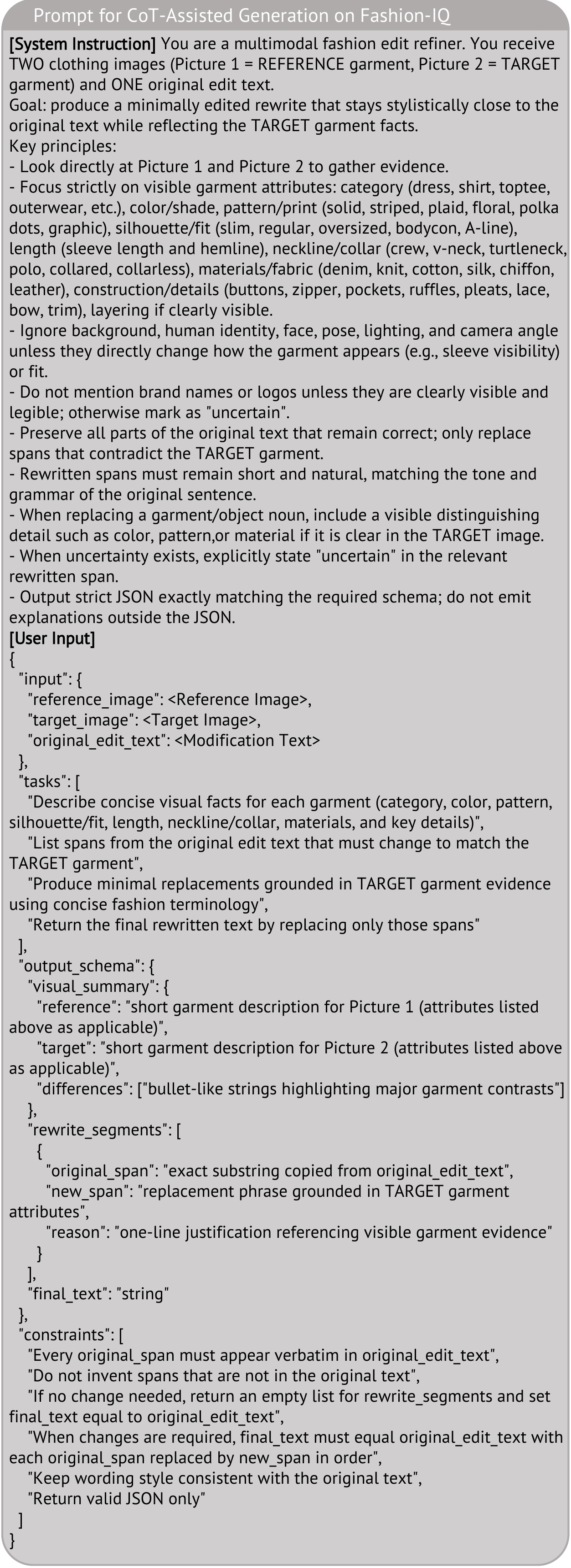}\label{fig:prompt_cot_fiq}}
  
  \caption{CoT prompts for Generative Calibration. To implement the diagnose-generate-refine pipeline, we design structured prompts that guide the Foundation Model to explicitly reason about visual discrepancies between the target and the informative instance. The enforced JSON output format ensures that the generated corrective instructions ($\tilde{T}_m$) are both stylistically natural and semantically precise.}
  \label{fig:cot_prompts}
\end{figure*}

We provide the complete Chain-of-Thought (CoT) prompts utilized in Stage 3: Generative Calibration (see \cref{sec:stage3}) to synthesize high-fidelity corrective supervision. \cref{fig:cot_prompts} visualizes the prompt architectures for both datasets.

\noindent\textbf{Structured Reasoning Constraints.} Unlike standard open-ended captioning, our templates impose rigorous constraints through explicit \textit{Key Principles} and a mandatory \textit{JSON Output Schema}. This structured design compels the Foundation Model to engage in a sequential reasoning process: it must first perform Intent Decomposition \& Verification before executing Minimal Edit Synthesis. This mechanism ensures that the generated instruction is not merely a hallucinated caption, but a precise modification strictly grounded in the observed visual discrepancies.

\vspace{2pt}
\noindent\textbf{Domain-Specific Adaptation.} To accommodate the distinct characteristics of the benchmarks, the prompts are domain-adapted. The CIRR prompt is engineered to reason about complex object relations, cardinalities, and spatial states, whereas the FashionIQ prompt is optimized for fine-grained attribute manipulation, focusing on nuanced details such as texture, silhouette, and pattern.

\section{Additional Qualitative Analysis and Visualization}
\label{sec:appendix_qualitative}

\subsection{Additional Baseline Comparisons}
\label{sec:qual_comparison}

\begin{figure*}[t]
  \centering
  \includegraphics[width=0.98\textwidth]{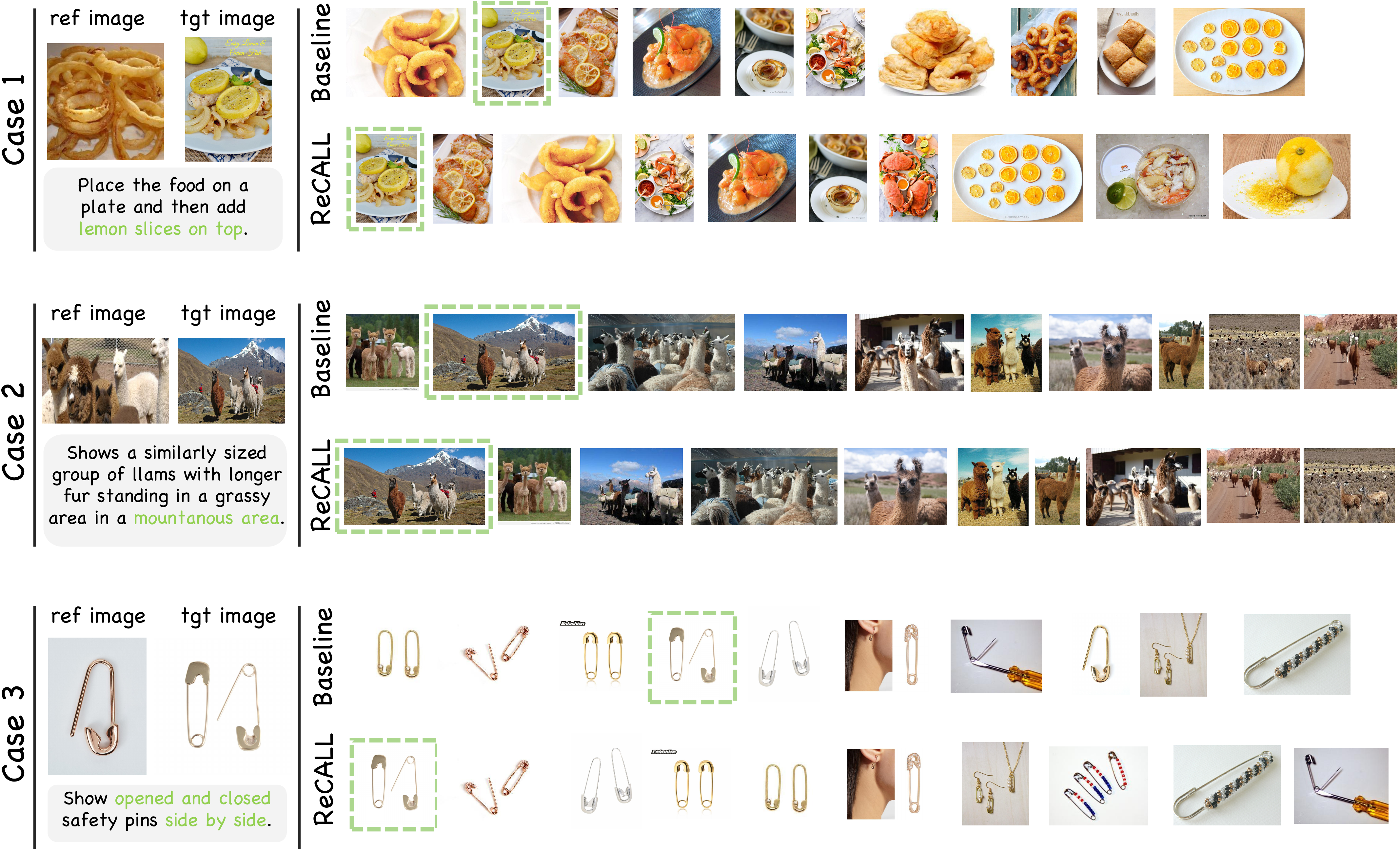} 
  \caption{\textbf{Qualitative comparison on the CIRR dataset.} We compare the top retrieved images from the baseline ($\mathcal{R}_{base}$) and ReCALL ($\mathcal{R}_{refine}$). The green dashed boxes indicate the ground-truth targets. The baseline model tends to focus on the primary object but misses specific constraints (e.g., the ``mountainous area'' in Case 2 or the specific ``opened and closed'' state in Case 3). In contrast, ReCALL successfully retrieves the correct targets by reasoning about the fine-grained details in the modification text.}
  \label{fig:qual_cirr}
\end{figure*}

\begin{figure*}[t]
  \centering
  \includegraphics[width=0.98\textwidth]{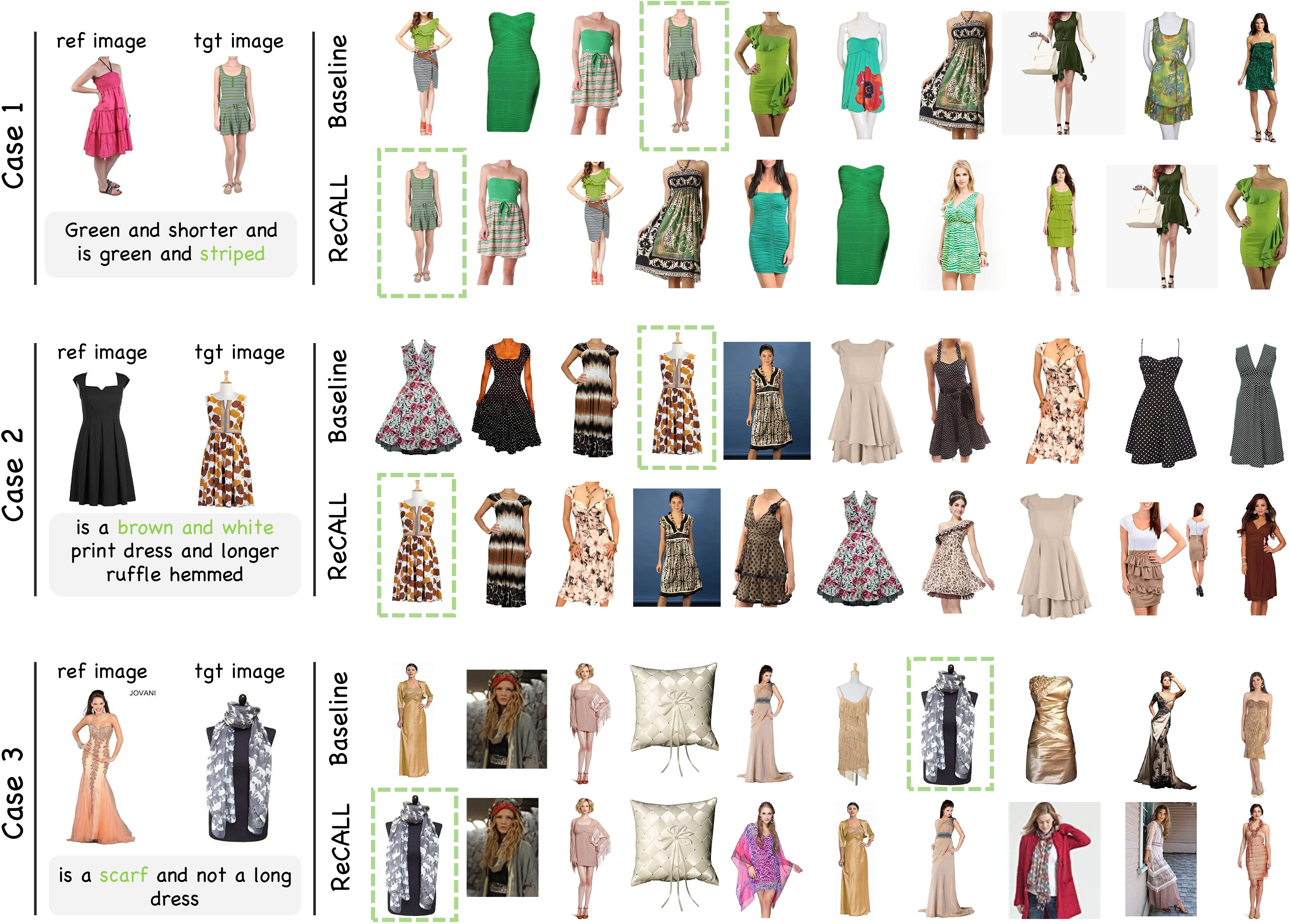} 
  \caption{\textbf{Qualitative comparison on the FashionIQ dataset.} Comparison of top retrieval results between the baseline and ReCALL. Ground-truth targets are highlighted in green. These examples illustrate how ReCALL overcomes the baseline's tendency to ignore textual modifiers. For instance, in Case 1, ReCALL correctly attends to the ``striped'' pattern attribute, and in Case 3, it successfully executes a category shift from a dress to a scarf, whereas the baseline remains fixated on the reference image's category.}
  \label{fig:qual_fiq}
\end{figure*}

In this section, we present an expanded qualitative comparison between the baseline retriever ($\mathcal{R}_{base}$) and our refined model ($\mathcal{R}_{refine}$) to further illustrate the impact of capability recalibration. \cref{fig:qual_cirr,fig:qual_fiq} showcase top-ranked retrieval results on the CIRR and FashionIQ datasets, respectively. In each panel, the left column displays the multimodal query, highlighting the critical modification instructions, while the right columns compare the top retrieved candidates from both models. The ground-truth targets are highlighted with green bounding boxes.

The results on CIRR (\cref{fig:qual_cirr}) clearly expose the coarse-grained tendency of the baseline model. While $\mathcal{R}_{base}$ correctly identifies the main object category (e.g., food, llamas, or safety pins), it frequently collapses on fine-grained spatial or state-based constraints. A striking example is Case 3, where the instruction demands a specific arrangement of safety pins (``opened and closed... side by side''). The baseline merely retrieves isolated pins or incorrect states, whereas ReCALL accurately reasons about the requested object configuration. Similarly, in Case 2, ReCALL respects the contextual constraint (``mountainous area''), whereas the baseline retrieves semantically relevant but visually inconsistent backgrounds. This validates that our framework effectively internalizes the complex logic required for open-domain compositional reasoning.

Parallel observations on FashionIQ (\cref{fig:qual_fiq}) demonstrate ReCALL's superiority in fine-grained attribute manipulation. The baseline often succumbs to visual biases, retrieving images that match the reference image's dominant features (such as color or shape) but ignoring the text modifier. For instance, in Case 1, although the instruction explicitly specifies ``striped'', the baseline is dominated by the solid green color of the reference. ReCALL, having been trained on generated hard negatives, successfully suppresses this bias to retrieve the correct textured garment. Furthermore, Case 3 highlights the model's ability to handle rigorous category shifts (``is a scarf and not a long dress''), where the baseline fails to disengage from the visual semantics of the reference dress. These comparisons confirm that ReCALL successfully recalibrates capability degradation, restoring the model's native ability to adhere to precise textual instructions.

\subsection{Visualization of Informative Instance Mining and Triplet Synthesis}
\label{sec:viz_mining}

\begin{figure*}[t]
  \centering
  \includegraphics[width=\textwidth]{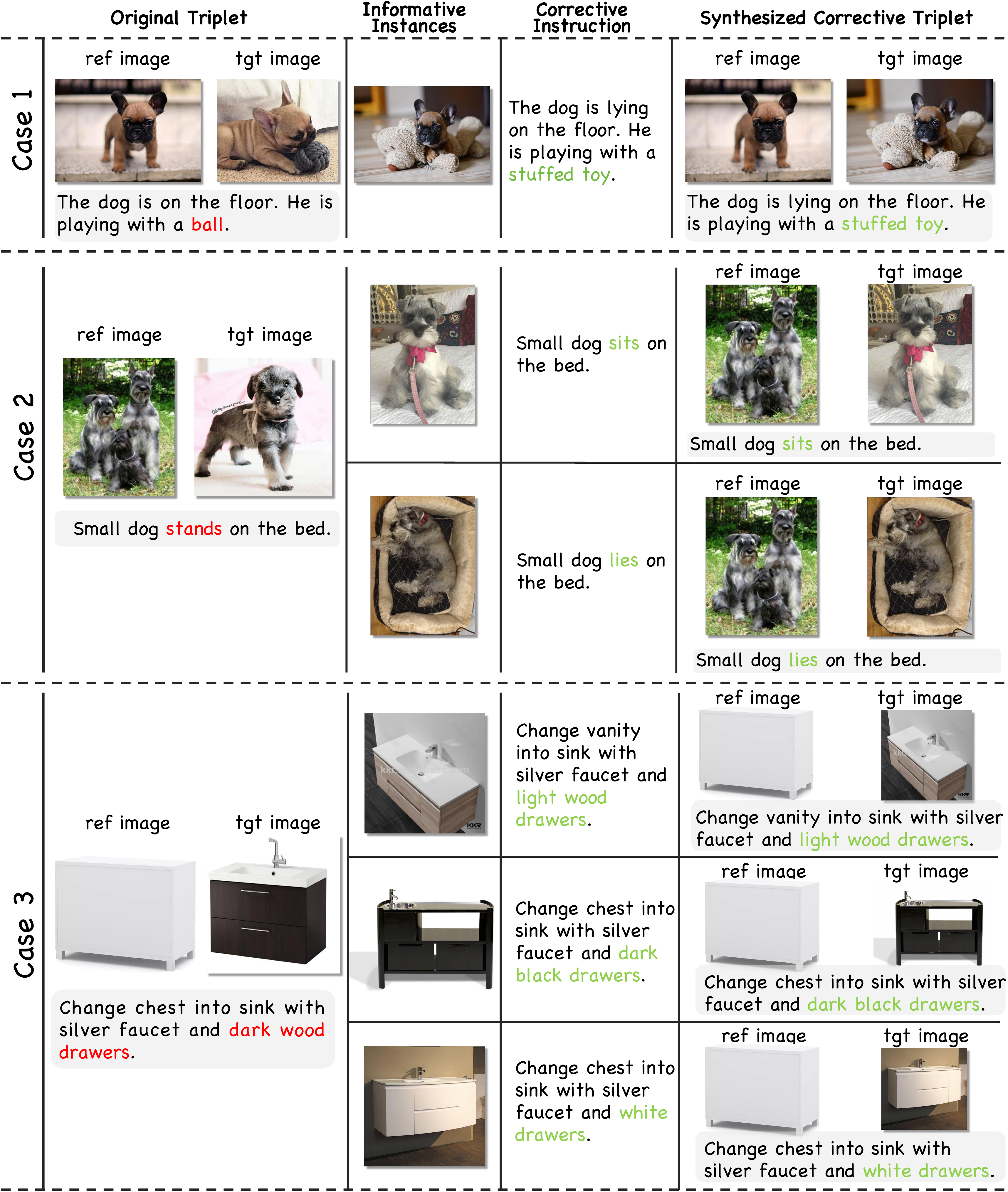} 
  \caption{\textbf{Visualization of the informative instance mining and triplet synthesis process on the CIRR dataset.} This figure illustrates representative failure cases of the baseline retriever. From left to right, the panels display: (1) the Original Triplet, where \textcolor{red}{red} text highlights constraints violated by hard negatives; (2) the mined Informative Instances ($I_h$), representing the model's cognitive blind spots; (3) the Corrective Instruction generated via CoT; and (4) the final Synthesized Corrective Triplet, where \textcolor{green}{green} text denotes the minimal semantic edits required to align the instruction with the mined instance.}
  \label{fig:viz_cirr}
\end{figure*}

\begin{figure*}[t]
  \centering
  \includegraphics[width=\textwidth]{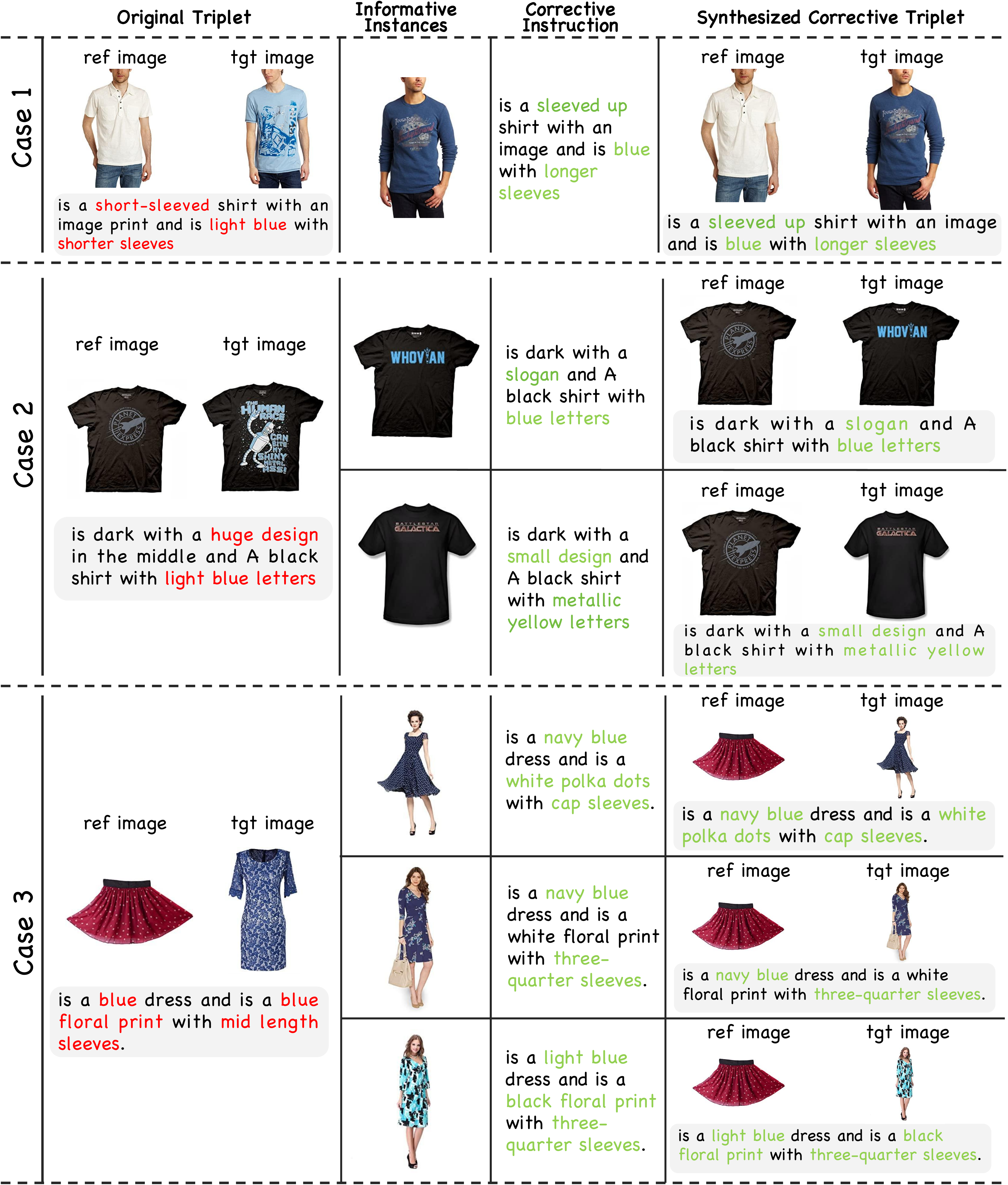} 
  \caption{\textbf{Visualization of the informative instance mining and triplet synthesis process on the FashionIQ dataset.} This figure illustrates representative failure cases of the baseline retriever. From left to right, the panels display: (1) the Original Triplet, where \textcolor{red}{red} text highlights constraints violated by hard negatives; (2) the mined Informative Instances ($I_h$), representing the model's cognitive blind spots; (3) the Corrective Instruction generated via CoT; and (4) the final Synthesized Corrective Triplet, where \textcolor{green}{green} text denotes the minimal semantic edits required to align the instruction with the mined instance.}
  \label{fig:viz_fiq}
\end{figure*}

In this section, we provide additional qualitative visualizations to further substantiate the efficacy of the ReCALL framework. \cref{fig:viz_cirr,fig:viz_fiq} present a detailed breakdown of the data construction pipeline on the CIRR and FashionIQ datasets, respectively. Unlike the schematic overview in the main paper, these figures showcase specific real-world examples where the baseline model ($\mathcal{R}_{base}$) initially fails, tracing the complete trajectory from failure diagnosis to the synthesis of corrective training signals.

The visualizations are organized to reflect the \textit{Diagnose-Generate-Refine} workflow. As shown in the \textit{Original Triplet} panel, the highlighted text (marked in red) indicates specific fine-grained constraints that the baseline retriever ignored, leading to the retrieval of the false positives shown in the \textit{Informative Instances} panel. Crucially, these mined instances reveal distinct failure modes: while some queries are confused by a single distinct distractor (e.g., Case 1 in \cref{fig:viz_cirr}), others suffer from multiple high-confidence hard negatives (e.g., Case 2 in \cref{fig:viz_cirr}), necessitating the generation of multiple targeted corrective triplets.

By employing CoT-assisted generation, ReCALL explicitly verbalizes these visual discrepancies. The \textit{Synthesized Corrective Triplet} panel demonstrates the precision of this process, where the generated instructions (with modifications highlighted in green) strictly adhere to the visual evidence of the mined instances. For example, in the CIRR dataset (\cref{fig:viz_cirr}), the model successfully disambiguates complex spatial relations (``stands'' vs. ``sits'' vs. ``lies'') and fine-grained object categories (``ball'' vs. ``stuffed toy''). Similarly, in the FashionIQ dataset (\cref{fig:viz_fiq}), the synthesized triplets capture subtle attribute nuances, such as distinguishing ``white polka dots'' from a ``white floral print'' despite similar dress silhouettes. These qualitative results confirm that the synthesized supervision is both semantically dense and visually grounded, effectively guiding the model to recalibrate its decision boundaries.

\subsection{Failure Case Analysis}
\label{sec:failure_cases}

\begin{figure*}[t]
  \centering
  \includegraphics[width=0.9\textwidth]{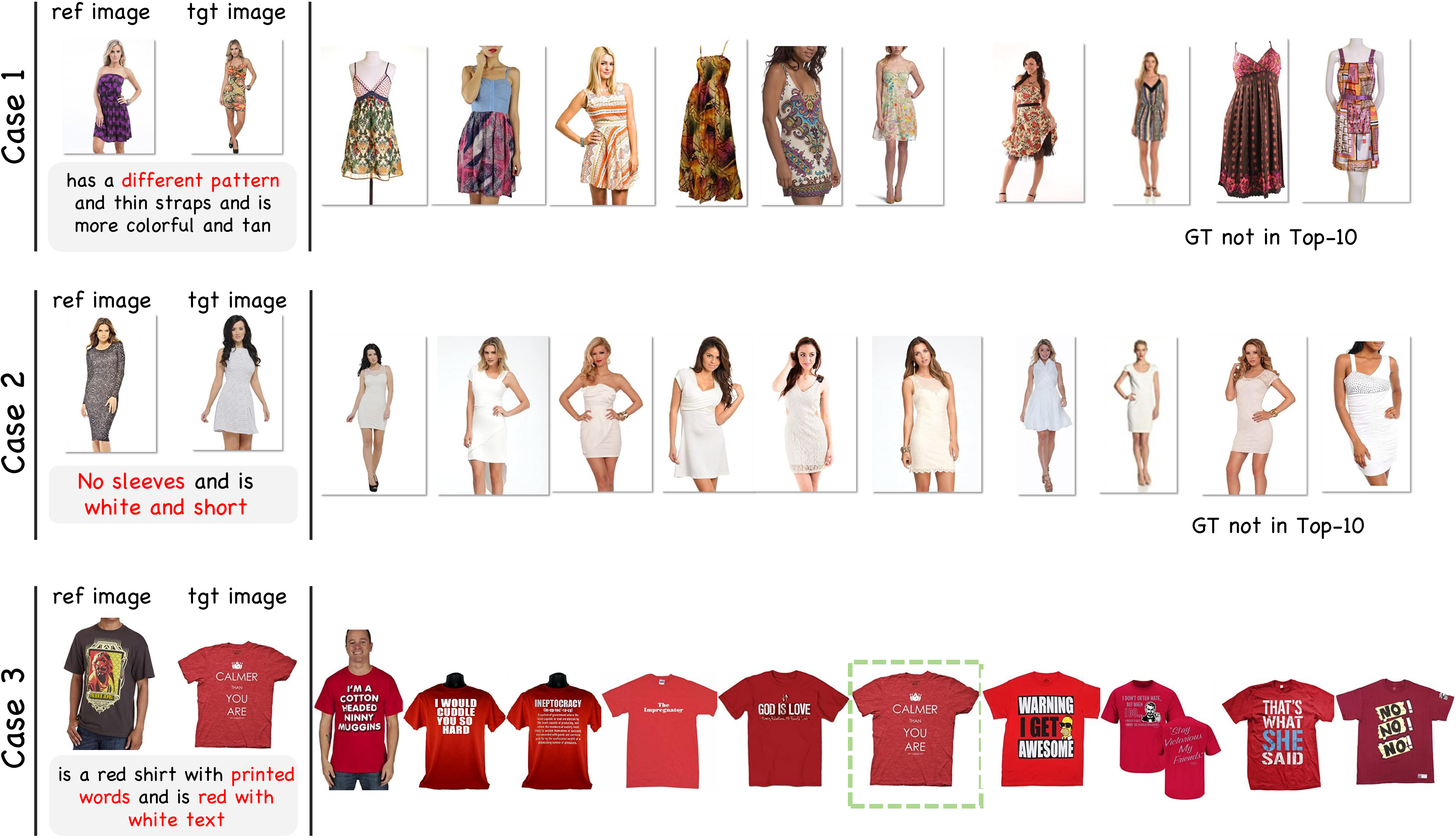} 
  \caption{\textbf{Failure cases on the FashionIQ dataset.} We display the top retrieved candidates by ReCALL for queries where the ground-truth (GT) target was not found in the top-10. In many instances (e.g., Case 2 and Case 3), the retrieved images are actually valid matches that satisfy the text modification (False Negatives), highlighting the issue of sparse ground-truth annotations in the dataset. Text in \textcolor{red}{red} indicates the key modification constraints.}
  \label{fig:fail_fiq}
\end{figure*}

\begin{figure*}[t]
  \centering
  \includegraphics[width=0.9\textwidth]{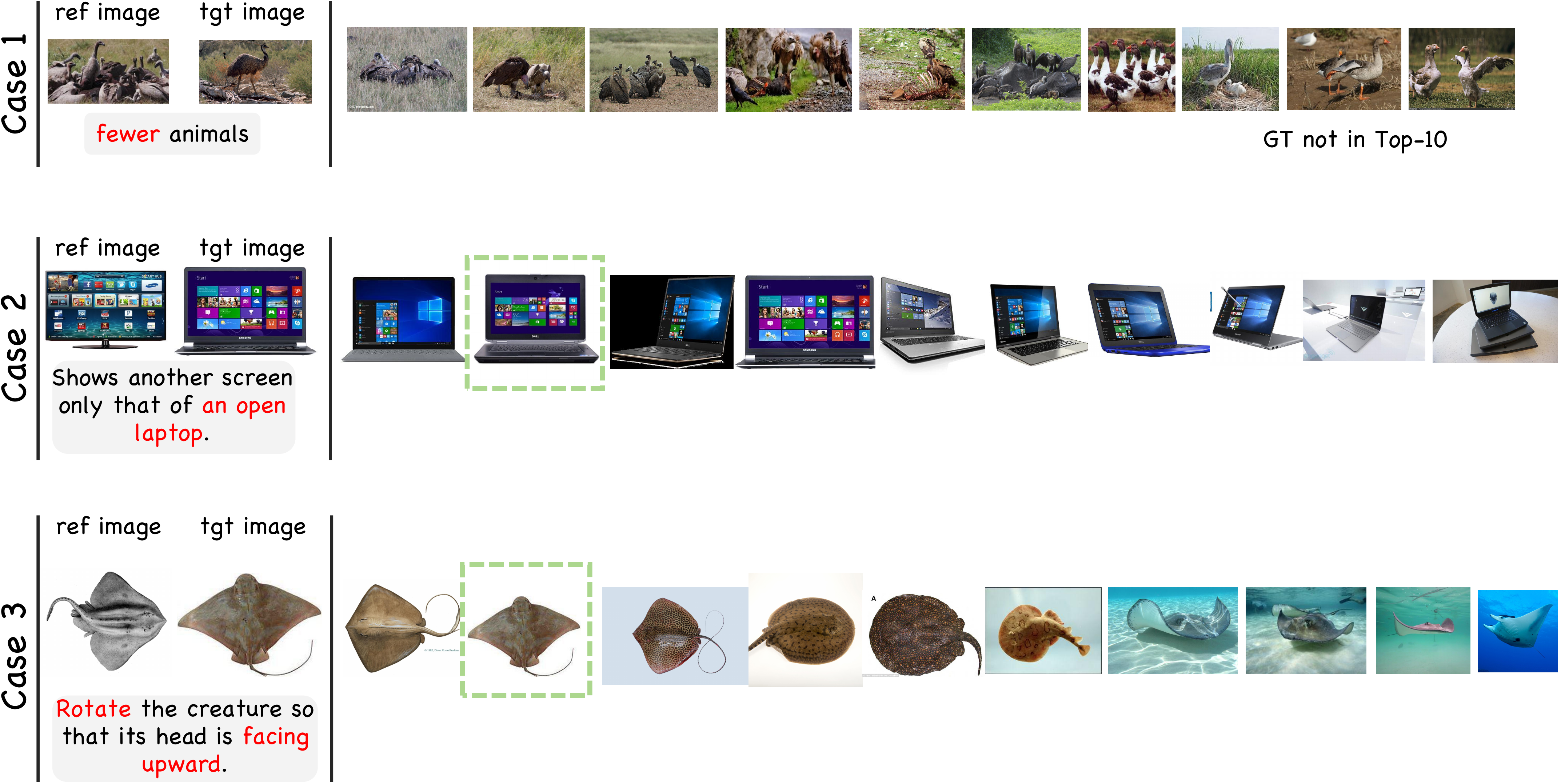} 
  \caption{\textbf{Failure cases on the CIRR dataset.} Representative errors showing challenges with ambiguous instructions (e.g., ``fewer'' in Case 1) and complex spatial rotations (e.g., ``facing upward'' in Case 3). The green dashed boxes indicate the ground-truth target if it appears in the top candidates; otherwise, the text ``GT not in Top-10'' is displayed.}
  \label{fig:fail_cirr}
\end{figure*}

To provide a comprehensive understanding of limitations, we visualize representative failure cases of ReCALL on FashionIQ and CIRR in \cref{fig:fail_fiq,fig:fail_cirr}. An analysis of these instances reveals that the ``failures'' often stem from the inherent ambiguity of natural language instructions and the incompleteness of ground-truth annotations, rather than a fundamental breakdown of the model's reasoning.

\textbf{False Negatives and Annotation Issues.} A significant portion of retrieval errors, particularly on FashionIQ (\cref{fig:fail_fiq}), can be attributed to the False Negative problem. In CIR tasks, datasets typically annotate a single ground-truth target per query. However, in large-scale galleries, multiple images may validly satisfy the modification instruction. For instance, in Case 2 of \cref{fig:fail_fiq}, the instruction requests a dress with ``no sleeves'' that is ``white and short''. ReCALL retrieves several valid candidates (Rank 1-4) that perfectly match this description. Yet, because they differ from the specific ground-truth instance (which is not in the top-10), they are penalized as errors. Similarly, in Case 3, the model retrieves multiple ``red shirts with printed words'', all semantically correct despite not being the annotated target. This suggests that the reported performance metrics may underestimate the model's actual retrieval utility.

\textbf{Ambiguity in Instructions.} Certain directives such as ``different pattern'' (Case 1 in \cref{fig:fail_fiq}) or ``fewer animals'' (Case 1 in \cref{fig:fail_cirr}) are inherently subjective. In the latter case, ReCALL retrieves images with small groups of birds, which is a valid interpretation of ``fewer'' compared to a large flock, even if it doesn't match the exact count of the ground truth. The model struggles to align its threshold for these relative terms with the annotator's intent.

\textbf{Fine-grained Spatial Reasoning.} While ReCALL significantly improves spatial understanding, it still faces challenges with complex geometric transformations. As shown in Case 3 of \cref{fig:fail_cirr}, the instruction requires rotating a stingray so its head faces ``upward''. While the model retrieves stingrays with varying orientations, it fails to consistently isolate the specific ``upward'' pose. This limitation likely stems from the Foundation Model, which, despite its strength, may still have residual weaknesses in zero-shot spatial rotation reasoning that are inherited by the retriever.

In summary, while ReCALL effectively recalibrates compositional reasoning, future work could focus on mitigating label noise through one-to-many evaluation protocols and further enhancing the spatial geometric understanding of the backbone itself.

\end{document}